\documentclass[%
 reprint,
 showkeys,
showpacs,%preprintnumbers,
 amsmath,amssymb,
 aps,
nofootinbib
]{revtex4-1}
\usepackage{graphicx}% Include figure files
\usepackage{dcolumn}% Align table columns on decimal point
\usepackage{bm}% bold math
\usepackage{hyperref}% add hypertext capabilities
\usepackage{float}
\usepackage{multirow}
\begin{document}
\preprint{APS/123-QED}
\setcounter{topnumber}{8}
\setcounter{bottomnumber}{8}
\setcounter{totalnumber}{8}
\title{Reconstruction of three-dimensional porous media \\ using generative adversarial neural networks}% Force line breaks with \\

\author{Lukas Mosser}
 \email{lukas.mosser15@imperial.ac.uk}
\author{Olivier Dubrule}%
 \email{o.dubrule@imperial.ac.uk}
\author{Martin J. Blunt}%
 \email{m.blunt@imperial.ac.uk}
\affiliation{%
 Department of Earth Science and Engineering, Imperial College London\\
}%

\date{\today}% It is always \today, today,

\begin{abstract}
To evaluate the variability of multi-phase flow properties of porous media at the pore scale, it is necessary to acquire a number of representative samples of the void-solid structure. While modern x-ray computer tomography has made it possible to extract three-dimensional images of the pore space, assessment of the variability in the inherent material properties is often experimentally not feasible.
We present a novel method to reconstruct the solid-void structure of porous media by applying a generative neural network that allows an implicit description of the probability distribution represented by three-dimensional image datasets. We show, by using an adversarial learning approach for neural networks, that this method of unsupervised learning is able to generate representative samples of porous media that honor their statistics. We successfully compare measures of pore morphology, such as the Euler characteristic, two-point statistics and directional single-phase permeability of synthetic realizations with the calculated properties of a bead pack, Berea sandstone, and Ketton limestone.
Results show that GANs can be used to reconstruct high-resolution three-dimensional images of porous media at different scales that are representative of the morphology of the images used to train the neural network. The fully convolutional nature of the trained neural network allows the generation of large samples while maintaining computational efficiency. Compared to classical stochastic methods of image reconstruction, the implicit representation of the learned data distribution can be stored and reused to generate multiple realizations of the pore structure very rapidly.

\end{abstract}

\maketitle

%\tableofcontents

\section{\label{sec:introduction}Introduction}
\subsection{\label{sec:image_reconstruction}Image Reconstruction}
The reconstruction and the evaluation of the material properties of porous media plays a key role across many engineering disciplines. Many physical processes such as the movement of multiple phases of fluids through sedimentary rocks are controlled by individual pores at the micron and sub-micron scale \cite{blunt_2017}.

In carbon capture and sequestration (CCS), the long term storage behavior is controlled by the physical and chemical interaction of super-critical $CO_2$ with the reservoir brine, as well as the spatial distribution and connectivity of minerals in the pore-space \cite{spiteri2005relative, kang2010pore}. The variability of the controlling properties such as the permeability of the host rock is determined by repeated experiments or numerical modeling of these processes. 

Using modern computer tomographic methods, it is possible to observe porous materials and evaluate their material properties at the micrometer scale (micro-CT) under static and transient conditions at high pressures and temperatures in near real time. Performing micro-CT imaging of porous media requires specialized, expensive equipment and in the case of CCS, only a single image of the investigated rock type is typically acquired. 

To evaluate the variability associated with the geometrical and mineralogical morphology of the pore-space, numerous physical experiments using the same rock type would have to be performed to obtain a distribution over larger volumes. Due to time and cost limitations inherent with the experimental acquisition of high-resolution images, this is often deemed unfeasible. Material properties governing the single and multi-phase flow behavior of porous media can be estimated from numerical solution of partial differential equations at the scale of a representative elementary volume (REV) and verified by experimental results \cite{Mostaghimi2013}.

Many sedimentary rocks consist of granular siliciclastic or carbonate materials. Boolean models use this fundamental characteristic of natural granular materials to emulate the shape of the arising pore space, due to an underlying random process that controls the distribution of the individual grains \cite{matheron1975random, serra1980boolean}. While for the classical Boolean model, the centers of the grains are uniformly distributed in space and grains can arbitrarily overlap, more complicated models with rigid hard sphere grains and more complex grain interaction functions have been developed \cite{MilleuxPoreux, arns2009boolean, rikvold1985, torquato2013random}. The framework of Boolean models also allows extension beyond spherical particles and enables derivation of the properties of material models as \newpage a function of the parameters of the underlying random process \cite{bretheau1989caracteristiques, Jeulin2000, lin1982cohen, yeong1998}.

In sedimentary rocks, the arrangement of individual grains occurs due to the transport of material from a high energy source to a low energy sink. Process models, where depositional mechanisms are simulated, have been shown to reproduce realistic granular reconstructions capturing the pore space morphology of granular sedimentary rocks \cite{Bakke2003}.

Spatial probabilistic models such as truncated Gaussian processes or sequential indicator simulation have been widely applied in the geosciences to model the spatial distribution of materials \cite{pyrcz2014geostatistical}. Many of these methods rely on two-point probability functions as a measure of spatial variability, whereas recent methods in geostatistics use training images as a basis for sample reconstruction \cite{caers2004multiple, mariethoz2010direct, meerschman2013practical}. These images are usually assumed to exhibit stationarity of the probability distribution of the properties of interest and rely on higher order multiple point statistics (MPS) to reconstruct stochastic random media. 

With MPS, the probability distributions are represented by training images and are sampled using a limited multi-scale neighborhood that captures the variation on a large scale, as well as fine structural details on smaller scales \cite{tahmasebi2014ms}. MPS based methods have been used in two and three-dimensional conditional simulation of spatial properties in reservoir-scale earth modeling applications \cite{comunian2011three}. The computational complexity of these methods is highly dependent on individual algorithms as well as the size of the domains used to sample from the training images \cite{mariethoz2014multiple}. Parallelized versions have been developed, reducing the computational time required to perform reconstruction using multiple point statistics \cite{straubhaar2011improved, huang2013gpu}.

Three-dimensional porous media have been reconstructed using a modified multiple-point statistics approach based on two-dimensional images of porous media \cite{Okabe2004, Okabe2005, Okabe2007}.

Stochastic methods based on simulated annealing allow the incorporation of arbitrary cost functions of statistical and morphological properties used in unconditional three-dimensional image reconstruction \cite{smith1983reconstruction, svergun1999restoring}. Recent advances have reduced the computational runtime of simulated annealing based methods for reconstruction of porous media, to the order of tens of hours per realization at the scale of $300^3$ voxels \cite{pant2016stochastic}. 

In the following section, we introduce a recently developed class of unsupervised machine learning methods called generative adversarial networks (GAN) that allow simulation of probability distributions given a set of training data \cite{goodfellow2014}. Volumetric generative adversarial networks have previously been applied to low-resolution three-dimensional CAD model synthesis, and practical applications of 3D-GANs are few compared to their two-dimensional counterpart \cite{wu2016}.  Integration of multi-resolution datasets incorporating image data across a number of length scales is possible in the GAN framework by using a Laplacian pyramid approach such as StackGAN \cite{2016arXiv161203242Z}.

We investigate the applicability of GANs to model three-dimensional textures of rocks based on three-dimensional binary representations of porous media acquired at the micrometer scale. We compare statistical, morphological and transport properties of the simulated images with those of the training images. We evaluate the single-phase directional permeability to show that the synthetic realizations sampled from the learned representation of the input data can capture single-phase flow properties of sedimentary rocks. 

Training of these neural networks involves finding a set of hyperparameters that lead to stable training \cite{goodfellow2016}. While this training can take on the order of tens of hours, the sampling of large volumetric domains occurs on the order of seconds on the current generation of graphical processing units (GPU). We show that in favorable cases convolutional neural networks incorporated in the GAN framework allow the generation of synthetic reconstructions of porous media that exceed the dimensions of their training images. Contrary to most existing simulation techniques the set of parameters used to generate synthetic realizations can be stored once trained allowing rapid generation of new samples to assess the variability of material properties.

While we apply GANs to a set of micro-CT images of porous media, the method can readily be applied to volumetric images of porous media obtained from other three-dimensional microscopy instruments such as nano or medical-CT instruments.

We discuss the challenges involved in training GANs for stochastic image reconstruction of porous media, as compared to other stochastic image reconstruction methods and evaluate the computational efficiency of GAN based image reconstruction. Finally, we provide empirical guidelines on the requirements of the input dataset to allow successful training of GANs on large three-dimensional voxel representations of natural porous media. 

All data used in this study is available in the public domain and we have made the code used for training, as well as example pre-trained models, available as additional supporting material \footnote{\url{https://github.com/LukasMosser/PorousMediaGan}}. A public dataset of high-resolution micro-CT images made available by the Imperial College Pore-Scale Modelling Group \footnote{\url{http://www.imperial.ac.uk/earth-science/research/research-groups/perm/research/pore-scale-modelling/micro-ct-images-and-networks/}}, of a spherical beadpack, Berea sandstone, and oolitic Ketton limestone will serve as benchmark cases to study the application of GANs to three-dimensional stochastic image reconstruction.
\begin{figure*}[t]
\includegraphics[keepaspectratio=True, width=\textwidth]{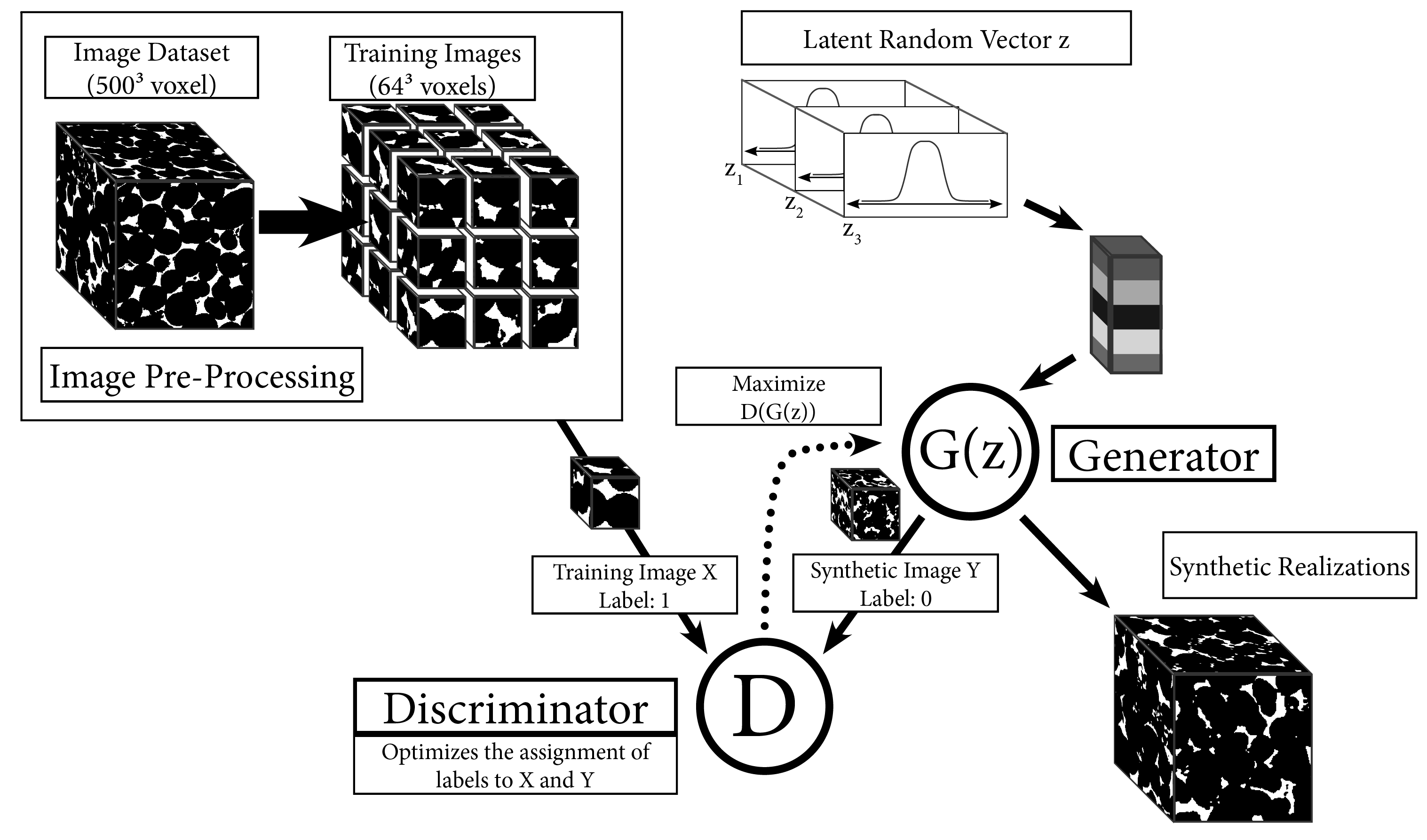}
\caption{\label{fig:GANoverview} Overview of the GAN training process. Segmented volumetric images are usually split into $64^3$ voxel training images. The generator G is a function that is applied to a sample from a latent random space $Z$ and creates a synthetic realization. We assume that samples drawn from the hidden latent space $Z$ are distributed according to a normal distribution (see Sec.~\ref{sec:gans}). The discriminator's role is to determine whether a sample is part of the training image dataset (label 1) or from the generator (label 0). The misclassification error is computed as a binary cross-entropy criterion and the error back-propagated to improve the discriminator's ability to distinguish real and "fake" images. Then the generator is updated to improve the quality of the produced samples and "fool" the discriminator. When sufficient image quality is obtained, training is stopped, and the discriminator may be discarded. The generator can now be used to create new samples. By providing larger latent vectors than used initially for training, larger output images can be produced.}
\end{figure*}

\section{\label{sec:gans}Generative Adversarial Networks}
In the following section, we present generative adversarial networks (GAN) in the context of three-dimensional image generation. 
Generative neural networks have been developed in the context of deep learning by Goodfellow et al. as a methodology to learn a representation of a high-dimensional probability distribution from a given dataset \cite{goodfellow2014}. In the context of image reconstruction, we refer to this dataset as a set of training images that present representative samples of the probability distribution underlying the image space.

GANs learn an implicit representation of the probability density as opposed to explicit density models. The main drawback of explicit density models is their computational cost which grows with the dimensionality of the samples and requires sequential simulation of each voxel. For high-dimensional samples such as volumetric image data, the computational cost is $O(N)$ where $N$ represents the number of voxels in the domain of interest and can easily exceed $10^9$ voxels for modern high-resolution micro-CT image data. Using any of these methods would make it intractable to generate a large number of very large samples. GANs have been designed to perform fast sampling from the learned density representation and allow full parallel generation, making them an ideal candidate to generate large volumetric images \cite{goodfellow2016}.

GANs consist of two differentiable functions: a discriminator $D$ and a generator $G$. The discriminator receives samples of the "real" dataset (Label 1) $x\thicksim p_{data}$ and "fake" samples $G(\mathbf{z})$ (Label 0) created by the generator from the hidden latent space $Z$ (see Fig.~\ref{fig:GANoverview} above). The latent space $Z$ is composed of independent real random variables, typically normally or uniformly distributed, that represent the random input to the generator $G$. The generator $G$  maps random variables from the latent space into the space of images. The discriminator's role is to assign a probability that a random sample is from the "real" data distribution $p_{data}$. The discriminator tries to label each sample correctly, while the generator tries to "fool" the discriminator into labeling the fake images as part of the true data distribution and therefore achieving $D(G(\mathbf{z}))$ close to one.

More formally we can define the loss i.e. the cost function for GANs as a minimization-maximization problem
\begin{equation}
\begin{aligned}
 \underset{G}{min} \underset{D}{max}\{\mathbb{E}_{\mathbf{x} \thicksim p_{data}(\mathbf{x})}[log(D(\mathbf{x}))]\\+\mathbb{E}_{\mathbf{z}\thicksim p_{\mathbf{z}}(\mathbf{z})}[log(1-D(G(\mathbf{z})))]\}\label{equ:gan_cost}
\end{aligned}
\end{equation}

Solutions to this optimization problem have been shown to be Nash equilibria, where each player achieves a local minimum of their loss function with respect to their parameters \cite{goodfellow2016}.

In practice we represent $G$ and $D$ by convolutional neural networks that are trained by a gradient descent based optimization method. Training is performed in two steps: First the discriminator is trained to maximize
\begin{equation}
\begin{aligned}
 J^{(D)}=\mathbb{E}_{\mathbf{x} \thicksim p_{data}(\mathbf{x})}[log(D(\mathbf{x}))]\\
 +\mathbb{E}_{\mathbf{z}\thicksim p_{\mathbf{z}}(\mathbf{z})}[log(1-D(G(\mathbf{z})))]\label{equ:discriminator_cost}
\end{aligned}
\end{equation}
while the parameters of the generator are fixed. This improves the ability of the discriminator to distinguish between real and fake images.

In a subsequent step we generate synthetic samples $G(\mathbf{z})$ by drawing samples $\mathbf{z}$ from an N-dimensional normal distributed latent space and train the generator to minimize
\begin{equation}
\begin{aligned}
J^{(G)}=\mathbb{E}_{\mathbf{z}\thicksim p_{\mathbf{z}}}[log(1-D(G(\mathbf{z})))]\label{equ:generator_cost}
 \end{aligned}
\end{equation}
while keeping the discriminator fixed. 

By minimizing Eq.~(\ref{equ:generator_cost}) the generator tries to "fool" the discriminator into believing that the samples $G(\mathbf{z})$ are real data samples. In this way the generator learns to represent a distribution $p_{g}(\mathbf{x})$ that is as close as possible to the real data distribution $p_{data}(\mathbf{x})$. When convergence is reached $p_{g}(\mathbf{x}) = p_{data}(\mathbf{x})$ and the value of the discriminator becomes $\frac{1}{2}$ as it cannot distinguish between the two anymore.

Initially, the discriminator $D$ outperforms the generator significantly making the gradient used to train the generator close to zero. Therefore, instead of minimizing $log(1-D(G(\mathbf{z}))$ for the generator, it is helpful to maximize $log(D(G(\mathbf{z}))$ \cite{goodfellow2016}.

GANs show highly unstable behavior during training and a large number of trial and error runs are required to find an optimal set of hyperparameters that allow stable training. A number of heuristics have been published which have been shown to stabilize GAN training, such as one-sided label smoothing and adding white noise to the input layer of the discriminator \cite{salimans2016improved, sonderby2016}. 

We provide a more detailed overview of the neural networks used in this study in Sec.~\ref{sec:network_architecture} and later provide suggestions on how to facilitate efficient training (see Sec.~\ref{sec:discussion}) for volumetric image datasets of porous media. 
%
%Methodology
%
\section{\label{sec:methodology}Methodology}
In the following section we outline the criteria used to evaluate the quality of simulations based on the training image datasets. We treat all images under the assumption of stationarity and the existence of a representative elementary volume.
%
%Evaluation Criteria
%
\subsection{\label{sec:evaluation_criteria}Evaluation Criteria}
%
%Two Point Statistics
%
\subsubsection{\label{sec:two_point_statistics}Two-Point Statistics}
\begin{figure}[h]
\includegraphics[keepaspectratio=True, scale=0.22]{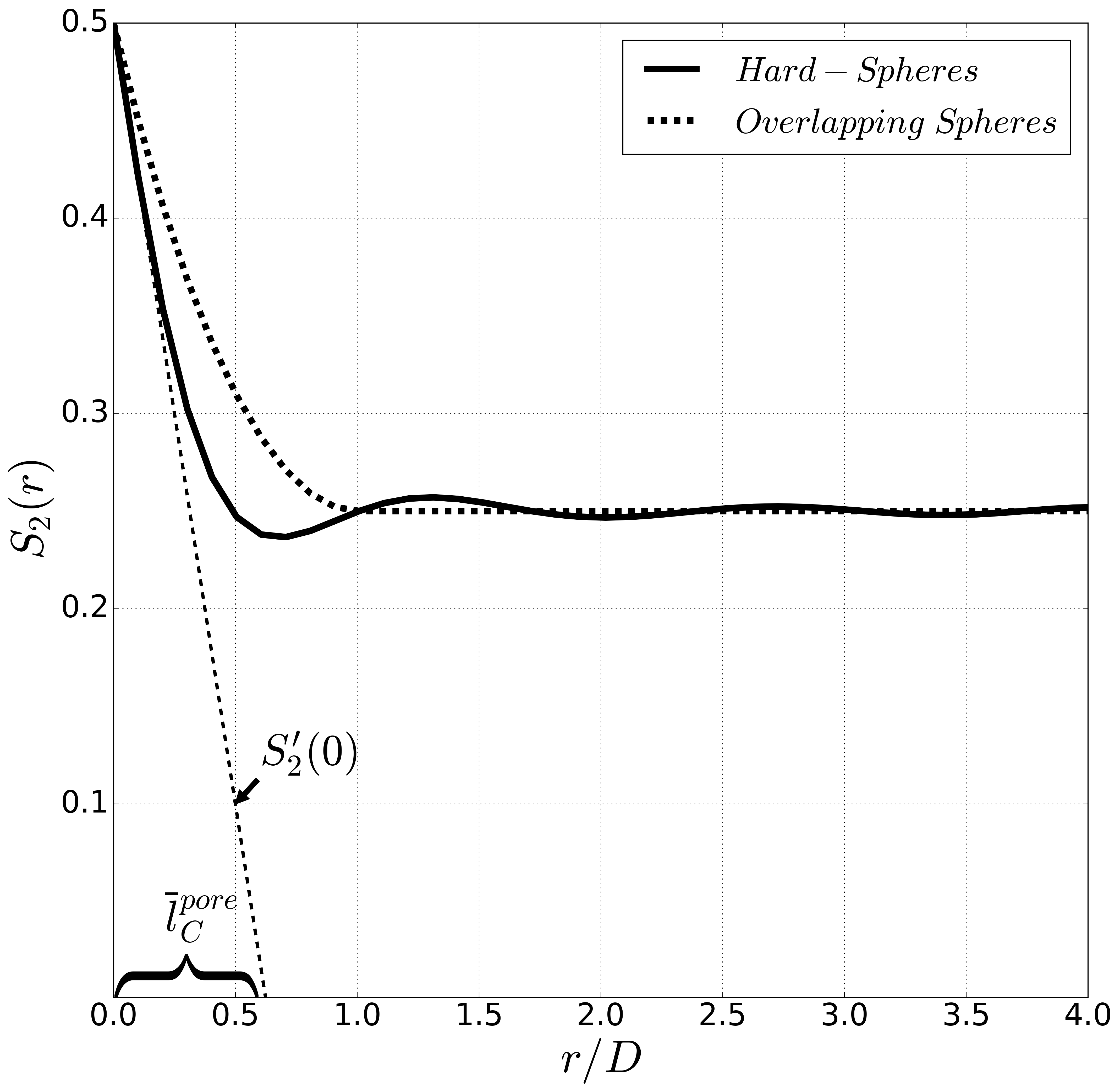}
\caption{\label{fig:hole_effect}Comparison of $S_2(r)$ for a Boolean model and a packing of hard spheres at a porosity $\phi=0.5$. $S_2$ exhibits exponential decay for the Boolean model, whereas a dampened oscillation is characteristic for packings of spheres. The mean chord length can be found at the intersection of the slope of $S_2$ at the origin with the x-axis [see Eq.~(\ref{equ:characteristic_pore_size})].}
\end{figure}

We characterize the second order structure of the porous media by calculating the two-point probability function of the pore phase. By assuming stationarity, this function is equivalent to the non-centered covariance \cite{MilleuxPoreux}:
\begin{equation}
S_2(\mathbf{r})=\mathbf{P}(\mathbf{x} \in P, \mathbf{x}+\mathbf{r} \in P) \ for \ \mathbf{x}, \mathbf{r} \in \mathbb{R}^d \label{equ:covariance}
\end{equation}
which is the probability $\mathbf{P}$ that two points $\mathbf{x}$ and $\mathbf{x}+\mathbf{r}$, separated by the lag vector $\mathbf{r}$, are located in the pore phase $P$. At the origin, $S_2(0)$ is equal to the porosity $\phi$. $S_2$ stabilizes around $\phi^2$ as $r\rightarrow\infty$ (Fig.~\ref{fig:hole_effect}).
Due to the anisotropic nature of many porous media, we compute $S_2(r)$ along the three Cartesian directions, as well as the radial average of $S_2(r)$.

\newpage
It is a well known result that the specific surface area $S_V$ of a porous medium can be expressed as a function of $S_2$ \cite{Debye1957}. In the case of an isotropic porous medium and in three-dimensions $S_V$ is related to $S_2$ by
\begin{equation}
S_V = -4S_2'(0) \label{equ:specific_surface_area}
\end{equation}
where $S_2'(0)$ is the derivative of $S_2(r)$ at the origin. 

Furthermore, the average chord length within the pore and the grain phases are \cite{torquato2013random}
\begin{subequations}
\label{equ:characteristic_pore_size}
\begin{eqnarray}
\overline{l}_c^{pore} = \frac{\phi}{S_2'(0)} \label{chord_length_pore}
\\
\overline{l}_c^{grain} = \frac{1-\phi}{S_2'(0)} \label{chord_length_grain}
\end{eqnarray}
\end{subequations}
which for the pore phase can be readily found from the intersection of the slope of $S_2(r)$ with the x-axis.

In favorable cases, it is possible to find analytical expressions of $S_2(r)$ from the spatial distribution and geometry of the grains.
A Boolean model of overlapping spherical grains of uniform spatial distribution exhibits an exponential decay of the covariance until the lag distance is equal to the diameter of the grains where it becomes zero \cite{MilleuxPoreux}. For porous media that can be well described by a Boolean model, we can estimate the size of the elementary Boolean grain from the decay of $S_2$.

Semi-analytical expressions for more complex models such as for a packing of hard spheres have been developed \cite{torquato1985characterisation}. Models of $S_2(r)$ for spherical packings exhibit a dampened oscillation. The shape of the estimated covariance, therefore, allows us to obtain information on the structure of the porous medium (see Fig.~\ref{fig:hole_effect} above).

The covariance $S_2(r)$ was estimated for the training images and the stochastic reconstructions generated by the trained GAN model.
For each GAN model, we evaluate the non-centered covariance $S_2$ as well as the specific surface area $S_V$ [Eq.~(\ref{equ:specific_surface_area})] and compare these to the values obtained from the original training images.

In our discussion on the required training image sizes (Sec.~\ref{sec:discussion}), we will use the average chord length and the specific surface area as possible indicators of the necessary training image size.

\subsubsection{\label{sec:morphological_measures}Morphological Measures}
It has been shown that flow properties at the pore-scale can be related to morphological characteristics of the void-solid interface of a porous medium \cite{scholz2015direct}. Hadwiger's theorem states that the size of a body in a $d$-dimensional space can be described by a linear combination of $d+1$ independent parameters characterizing the body. In three dimensions we can, therefore, define four so-called Minkowski functionals that fully characterize the size of a three-dimensional object. We compute estimates of three Minkowski functionals; the porosity $\phi$, the specific surface area $S_V$ and the Euler characteristic $\chi_V$ corresponding to the zero, first and 3\textsuperscript{rd} order functionals. We compute the densities of the Minkowski functionals by dividing by the volume $V$.

The Minkowski functional of order zero is the porosity, defined as the ratio of volume of the void space to the bulk volume of the sample
\begin{equation}
\phi = \frac{V_{pore}}{V} \label{equ:porosity}
\end{equation}
and is, therefore, a measure of the ability of a porous medium to store fluids.

The Minkowski functional of rank one is the specific surface area $S_V$. 
\begin{equation}
S_V = \frac{1}{V}\int{dS}\label{equ:specific_surface_area_minkowski}
\end{equation}
where integration occurs over the void-solid interface S. The specific surface area $S_V$ has dimensions of $\frac{1}{length}$ 
and its inverse allows us to define a characteristic pore size.

The specific Euler characteristic is closely related to the order three Minkowski functional and represents a dimensionless quantity defined as 
\begin{equation}
\chi_v = \frac{1}{4 \pi V}\int{\frac{1}{r_1 r_2}dS}\label{equ:specific_euler_characteristic_minkowski}
\end{equation}
where $r_1$ and $r_2$ are the principal radii of curvature of the void-solid interface.
To compute $\chi_V$ we do not directly evaluate the integral in Eq.~(\ref{equ:specific_euler_characteristic_minkowski})
but instead make use of a relationship for the Euler characteristic of arbitrary polyhedra,
\begin{equation}
\chi = V-E+F-O \label{equ:euler_characteristic_polyhedra}
\end{equation}
where $V$ is the number of vertices, $E$ the number of edges, $F$ the number of faces and $O$ the number of objects \cite{blunt_2017}.
This expression is the basis for efficient algorithms to compute Minkowski functionals of arbitrary geometric bodies represented as volumetric voxelized domains \cite{Lang2001}. To compute these three Minkowski functionals we have used the open-source image morphological software library MorphoLibJ \cite{legland2016morpholibj}.

While the porosity expresses the ability to store fluids in a porous medium, adsorption and dissolution processes are controlled by the specific surface area. The Euler characteristic allows the connectivity of the porous medium to be characterized, which is a critical component in the ability of fluids to flow. Reconstructions of porous media should therefore closely match the observed Minkowski functionals to represent the behavior of relevant physical processes at the pore-scale.

The direct computation of the specific surface area $S_V$ and porosity $\phi$ from images allows us to perform a comparison with the values obtained from estimates obtained by computing the empirical non-centered covariance $S_2(r)$ [see Eq.~(\ref{equ:specific_surface_area})].

\begin{table*}
\centering
\squeezetable
\caption{Neural network configurations and hyperparameters used to train on voxelized image subsets.}
\label{tab:dcgan}
\begin{ruledtabular}
\begin{tabular}{cccc}
                                                 & \multicolumn{3}{c}{Training Image Dataset}                                                                                                       \\
                                                 & Beadpack                                              & Berea                                                 & Ketton                           \\ \hline
\multicolumn{1}{c|}{Training Image Size}         & \multicolumn{1}{c|}{$128^3$ voxels}                   & \multicolumn{1}{c|}{$64^3$ voxels}                    & $64^3$ voxels                    \\ \hline
\multicolumn{1}{c|}{Latent Space $\mathbf{z}$ Dimension}  & \multicolumn{1}{c|}{100}                              & \multicolumn{1}{c|}{512}                              & 100                              \\ \hline
\multicolumn{1}{c|}{Generator Filters $N_G$}     & \multicolumn{1}{c|}{64}                               & \multicolumn{1}{c|}{64}                               & 64                               \\ \hline
\multicolumn{1}{c|}{Discriminator Filters $N_D$} & \multicolumn{1}{c|}{8}                                & \multicolumn{1}{c|}{16}                               & 16                               \\ \hline
\multicolumn{1}{c|}{Optimizer}                   & \multicolumn{3}{c|}{Generator + Discriminator: Adam}                                                                                             \\ \hline
\multicolumn{1}{c|}{Learning Rate / Momentum}    & \multicolumn{1}{c|}{$2\times10^{-4}$ / 0.5}           & \multicolumn{1}{c|}{$2\times10^{-4}$ / 0.5}           & $2\times10^{-4}$ / 0.5           \\ \hline
\multicolumn{1}{c|}{Stabilization}               & \multicolumn{1}{c|}{White Noise ($\sigma \ = \ 0.1$)} & \multicolumn{1}{c|}{Label Smoothing ($\epsilon=0.1$)} & White Noise ($\sigma \ = \ 0.1$)
\end{tabular}
\end{ruledtabular}
\end{table*}
\newpage
\subsubsection{\label{sec:single_phase_permeability}Single-Phase Permeability}
To evaluate the single-phase permeability of the porous media and their generated synthetic reconstructions we solve the Stokes equations for slow, incompressible flow assuming small inertial forces. 
\begin{subequations}
\label{equ:stokes}
\begin{eqnarray}
\nabla \cdot \mathbf{v}&=&0\label{equationa}
\\
\mu \nabla^{2} \mathbf{v}&=&\nabla p\label{equationb}
\end{eqnarray}
\end{subequations}
The Stokes equations are solved on the domain that is connected to the fluid inlet and outlet. This allows us to define an effective porosity where only the fraction of the pore space that also contributes to flow is considered
\begin{equation}
\phi_{eff} = \frac{V_{flow}}{V} \label{equ:effective_porosity}
\end{equation}

A finite difference method to solve Eq.~(\ref{equationa})-(\ref{equationb}) on pore-space representations has been implemented as a parallel flow solver, in the free open source numerical framework OpenFOAM \cite{weller1998, Mostaghimi2013}. 

\subsection{\label{sec:network_architecture}Neural Network Architecture}
\begin{figure*}
\includegraphics[keepaspectratio=True, width=\textwidth]{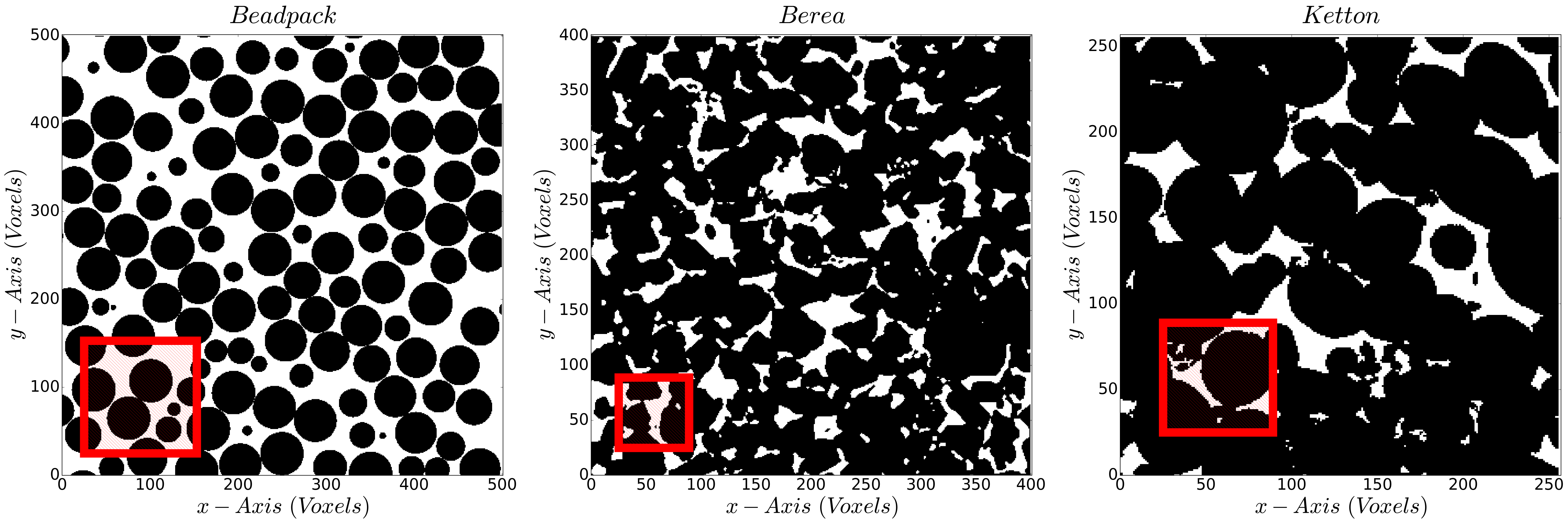}
\caption{\label{fig:cross_sections_overview} Cross-sections of the three binary images that have been reconstructed.
The bordered regions indicate the size of the training images extracted from the full dataset. The beadpack consists of spheres of equal diameter ($d=50 \ voxels$). The Berea sandstone is an angular granular sandstone that shows traces of dispersed clay. The oolitic Ketton limestone consists of ellipsoidal grains showing inter and intra-granular porosity. The voxel sizes are 3 $\mu m$ for the bead pack as well as the Berea sandstone and 15.2 $\mu m$ for the Ketton sample.}
\end{figure*}
The neural network architecture used for the three-dimensional image reconstruction corresponds to a volumetric version of the DCGAN network \cite{Radford2016}. The network consists of two independent fully convolutional neural networks, the generator $G$ and the discriminator $D$. Upsampling from the input latent vector $\mathbf{z}$ is performed by volumetric transposed convolution, followed by batch normalization and a rectified linear unit (ReLU) activation function in all layers except the last \cite{ioffe2015, nair2010rectified}. 

The discriminator $D$ receives images sampled from the latent space by the generator $G(\mathbf{z})$ and images from the set of training images representing $p_{data}(\mathbf{x})$. Therefore, the size of the input layer of the discriminator corresponds to the dimensions of the input training images. The discriminator consists of volumetric convolution layers combined with LeakyReLU activation functions \cite{Maas2013}. The final convolutional layer of the discriminator is followed by a Tanh activation function.

This combination of generator and discriminator neural network architectures has previously been applied to subsets of the Imagenet and CIFAR-10 datasets \cite{Radford2016}. The hyperparameters for the generator to be used in the optimization of the neural network architecture are the number of trainable convolutional filters in each layer of the neural network $N_{G, F}$, $N_{D, F}$ and the size of the latent vector $\mathbf{z}$. 

The generator and discriminator are optimized using a gradient descent based method where the parameters $w$ are changed by taking $k$ steps in the gradient
\begin{equation}
w^{k+1}=w^k-\alpha \nabla{f(w^k)} \label{equ:gradient_descent}
\end{equation}
where $\alpha$ is the learning rate. We have used the gradient descent based optimiser ADAM for optimization of both the generator and discriminator \cite{kingma2014adam}. 

GANs have been shown to exhibit unstable behavior during training. The addition of Gaussian noise to the input of the discriminator provides an effective measure to prevent mode collapse and stabilize the training process \cite{sonderby2016}. An additional stabilization measure called one-sided label smoothing, wherein the class label of 1 for real images is replaced by a new value of $1-\epsilon$ has been empirically shown to improve training of GANs \cite{salimans2016improved}.

Both label smoothing and white noise addition to the input of the discriminator have been used in this study to stabilize the training based on the volumetric image datasets. Table \ref{tab:dcgan} gives an overview of the neural network hyperparameters used for each evaluated sample, the hyperparameters and the stabilization measure used during training.

Images generated by the GAN were post-processed using a $3^3$ median filter to remove single-pixel noise. The resulting images are grayscale images with all voxel values close to zero or one. To compare the resulting images to the binary training images, we segment the generated images using Otsu's method \cite{otsu1975threshold}.
\newpage
\section{\label{sec:experimental_data}Experimental Data}

\subsection{\label{sec:microct_data_processing}Image Data and Processing}
To evaluate the applicability of GANs for reconstruction of natural porous media we use three previously acquired datasets. All images have been segmented into a three-dimensional binary voxel representation of the pore space (white) and grain structure (black) (Fig.~\ref{fig:cross_sections_overview}).
We create a training database of images by extracting sub-volumes from the voxelized binary images. Ideally, these training images should represent independent domains, but due to the limited size of these images, we extract subsets that overlap. 

Training image sizes were chosen based on an estimate of the average grain size for each sample. To be able to match the covariance $S_2(r)$ [Eq.~(\ref{equ:covariance})] and image morphological characteristics, training images larger than the structuring element were necessary. We discuss this requirement in more detail in the discussion of our results (see Section~\ref{sec:discussion}). Due to computational limitations, training image sizes exceeding $128^3$ voxels were not considered.
\subsubsection{\label{sec:description_beadpack}Beadpack}
The beadpack is based on a real packing of equally sized ceramic grains in a disordered close packing \cite{finney1970random}. The image consists of $500^3$ voxels with a size of 3 $\mu m$. The size of an individual sphere is 50 voxels. 1727 training images were extracted of size $128^3$ voxels corresponding to a spacing of 32 voxels between them in the original image. 
\subsubsection{\label{sec:description_berea}Berea}
Berea sandstone is a fluvial sandstone of medium to fine grain size (Wentworth classification) \cite{Survey}.
The individual grains are bonded by clays. The sample analyzed in this study was acquired from an outcropping of the Berea sandstone in a quarry near Berea, Ohio. De Witt showed that the Berea sandstone was deposited in the early Carboniferous (354-323 Mya) \cite{Stratigraphy}.

The image of Berea sandstone consists of angular grains with no clay presence in the intergranular pore-space. The image has dimensions of $400^3$ voxels with a voxel size of 3 $\mu m$.

To capture the local interaction of grains we have extracted training images at $64^3$ voxels which allows a number of grains to be present in one training image (see Sec.~\ref{sec:discussion}). Due to the small image size of $400^3$ voxels, subvolumes were extracted at a spacing of 16 voxels. In all, 10647 training images were used for the image reconstruction.
\begin{figure*}[ht]
\includegraphics[keepaspectratio=True,width=\textwidth]{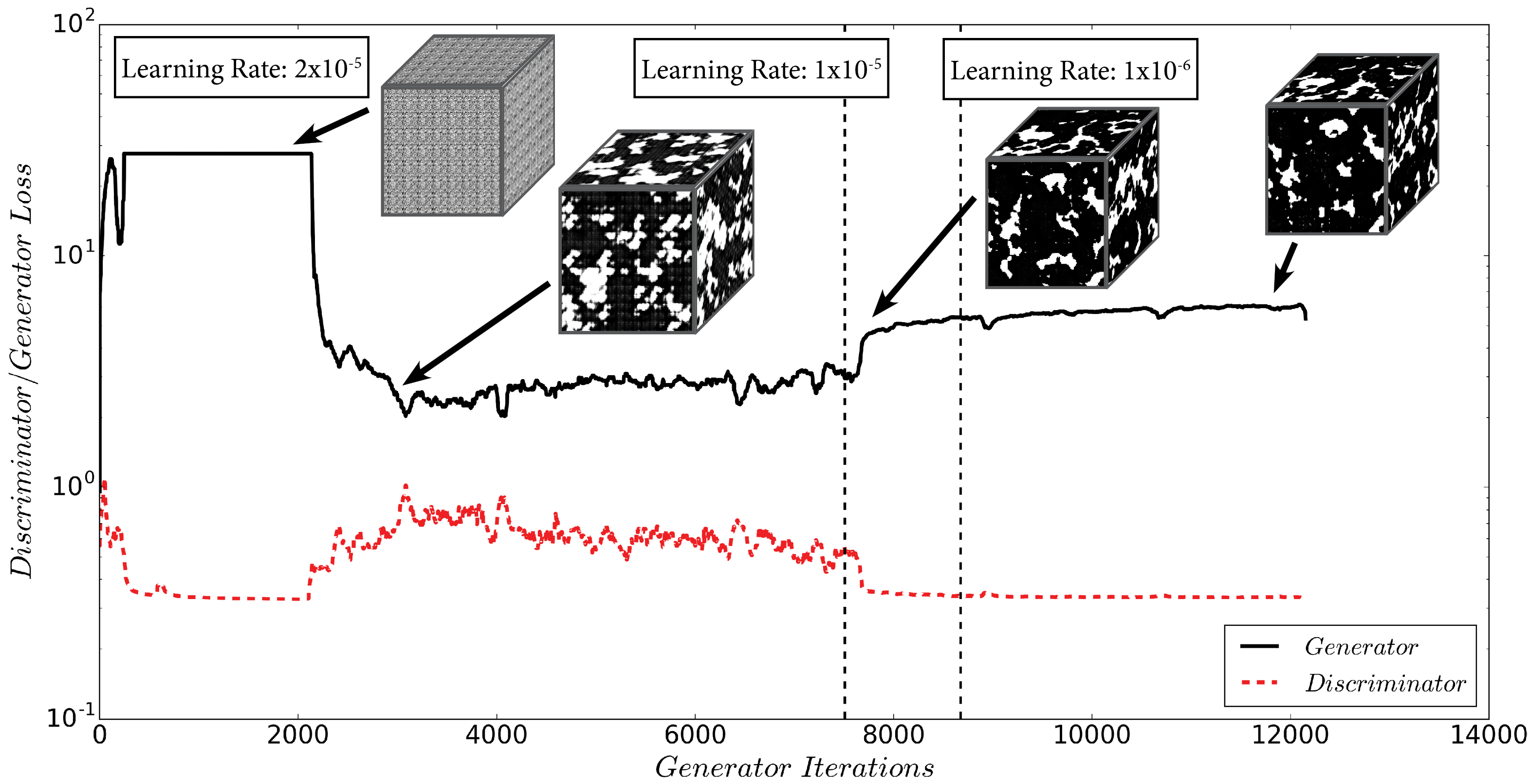}
\caption{\label{fig:berea_training_curve} Value of the cost function i.e. loss of the discriminator and generator [Eq.~(\ref{equ:discriminator_cost})-(\ref{equ:generator_cost})] for the GAN trained on the Berea sandstone. The samples shown were computed with the same random number seed, showing the evolution of a single realization during training. Initially, image quality is very low and random noise can be observed. After 2000 generator iterations a drop in the generator loss function is observed and coarse structures can be identified in the resulting sample. Loss functions in GAN models do not reflect improvement in image quality which can be observed from samples. Learning rates [see Eq.~(\ref{equ:gradient_descent})] were reduced after sufficient image quality was reached and training stopped based manual inspection of Minkowski functionals and two-point statistics.}
\end{figure*}
\subsubsection{\label{sec:description_ketton}Ketton}
The Ketton sample is an oolitic limestone of Jurassic age (201.3-145 Mya). The sample was acquired from a quarry of Lincolnshire limestone in the North-East of England. The oolites contained in the Lincolnshire formation are mainly non-ferroan calcite grains. The oolitic limestones of the Lincolnshire show a wide variety of cementation, ranging from uncemented oolite sands with no intergranular cement to heavily ferroan spar-cemented oolites with infilled microporosity \cite{emeryorigin}. Microstructures in the pore space can be observed that lead to a reduction in porosity (Fig.~\ref{fig:cross_sections_overview}).

The Ketton sample chosen for this study consists of large grains compared to the overall image size. The image used for the following evaluation has been downsampled from a $500^3$ voxel representation to an image size of $256^3$ voxels. This allows more grains to be resolved per training image extracted from the full volume. The downsampled voxel size is 15.2 $\mu m$. 

Training images were extracted at a sub-volume size of $64^3$ with a spacing of 8 voxels leading to a total of 15624 training images. The small spacing of the training images results from the small CT image size of $256^3$ voxels.
\begin{table*}[b]
\squeezetable
\caption{Chord lengths $\overline{l}_C$ for the pore and grain phase [Eq.~(\ref{equ:characteristic_pore_size})] determined from the radial averaged covariance $S_2(r)$ of each training image and corresponding realizations generated by the GAN model.  The specific surface area $S_V$ and porosity $\phi$ were evaluated for each of the samples using direct image morphological computation and derived from the covariance. Close agreement between estimates of the porosity and specific surface area can be observed for values determined by direct image morphological estimation and derived values obtained from the radial averaged covariance.}
\label{tab:pore_size_comparison}
\begin{ruledtabular}
\begin{tabular}{ccccccccccccc}
                                                          & \multicolumn{4}{c}{Beadpack}                                                          & \multicolumn{4}{c}{Berea}                                                             & \multicolumn{4}{c}{Ketton}                                                \\ \hline
\multicolumn{1}{c|}{$\overline{l}_C^{pore}$ {[}voxel{]}}  & \multicolumn{4}{c|}{20}                                                               & \multicolumn{4}{c|}{10}                                                               & \multicolumn{4}{c}{9}                                                     \\
\multicolumn{1}{c|}{$\overline{l}_C^{grain}$ {[}voxel{]}} & \multicolumn{4}{c|}{36}                                                               & \multicolumn{4}{c|}{41}                                                               & \multicolumn{4}{c}{64}                                                    \\ \hline
\multicolumn{1}{c|}{}                                     & \multicolumn{2}{c|}{Training Image}       & \multicolumn{2}{c|}{Synthetic}            & \multicolumn{2}{c|}{Training Image}       & \multicolumn{2}{c|}{Synthetic}            & \multicolumn{2}{c|}{Training Image}       & \multicolumn{2}{c}{Synthetic} \\
\multicolumn{1}{c|}{Minkowski Functional}                 & $S_2(r)$ & \multicolumn{1}{c|}{Direct} & $S_2(r)$ & \multicolumn{1}{c|}{Direct} & $S_2(r)$ & \multicolumn{1}{c|}{Direct} & $S_2(r)$ & \multicolumn{1}{c|}{Direct} & $S_2(r)$ & \multicolumn{1}{c|}{Direct} & $S_2(r)$      & Direct     \\ \hline
\multicolumn{1}{c|}{Porosity $\phi$}                      & 0.363       & \multicolumn{1}{c|}{0.359}  & 0.368       & \multicolumn{1}{c|}{0.366}  & 0.196       & \multicolumn{1}{c|}{0.198}  & 0.199       & \multicolumn{1}{c|}{0.197}  & 0.127       & \multicolumn{1}{c|}{0.119}  & 0.119            & 0.119      \\
\multicolumn{1}{c|}{$S_v \times 10^{-2} \ [\frac{1}{voxel}]$}            & $7.0$      & \multicolumn{1}{c|}{$7.3$} & $6.9$      & \multicolumn{1}{c|}{$7.5$} & $7.5$      & \multicolumn{1}{c|}{$8.2$} & $7.9$      & \multicolumn{1}{c|}{$8.5$} & $5.2$      & \multicolumn{1}{c|}{$5.2$} & $4.7$           & $5.2$   
\end{tabular}
\end{ruledtabular}
\end{table*}
\section{\label{sec:results}Results}
Three GANs were trained based on the network architectures highlighted in Sec.~\ref{sec:network_architecture}. The training time for each dataset was 24 hours. Manual inspection of synthetic realizations was performed during training to ensure convergence and intermediate evaluation of the covariance and Minkowski functionals.

Figure \ref{fig:berea_training_curve} shows the training curve for the Berea sandstone dataset. Initially the generator loss function [see Eq.~(\ref{equ:generator_cost})] is very high and no structural components can be observed in the samples. After a large reduction in the loss function of the generator, initial structures are observed. Image reconstruction quality significantly improves with the number of generator iterations, but cannot be linked to the loss function of the generator. This can be observed from the increase in generator loss at the end of training while image quality improves significantly.

The final GAN models were subsequently evaluated in terms of their directional and radial averaged non-centered covariance $S_2(r)$, Minkowski functionals and the single-phase permeability.

For all datasets, 20 realizations were generated using the trained GAN model. In the following section, we present the results of the evaluation of the properties outlined in Sec.~\ref{sec:evaluation_criteria} and compare these to the properties of the original input training image.
\subsection{\label{sec:results_beadpack}Beadpack}
\begin{figure}
\includegraphics[keepaspectratio=True, width=\columnwidth]{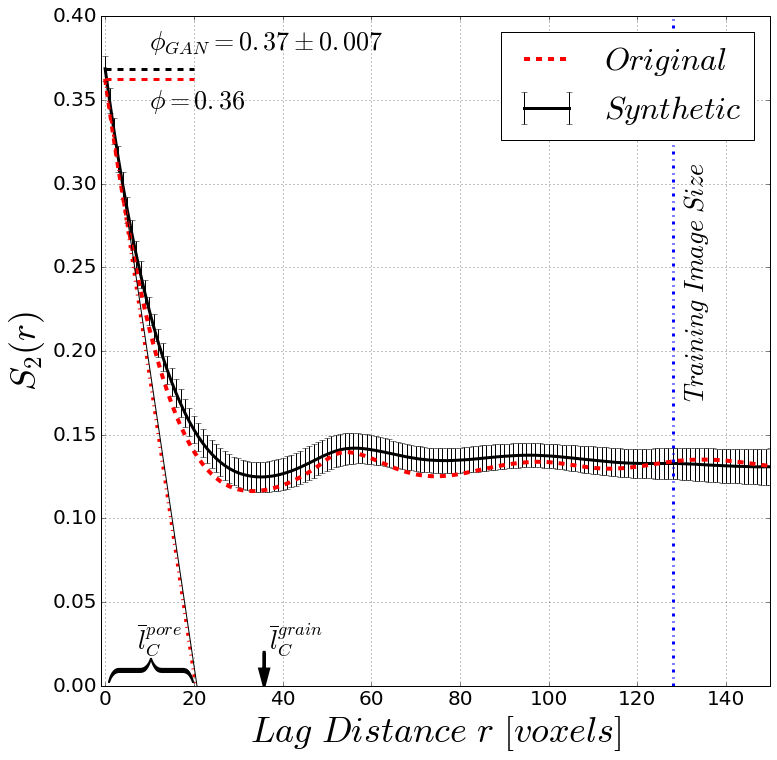}
\caption{\label{fig:radial_correlation_beadpack} Radial averaged covariance $S_2(r)$ for the beadpack sample and 20 synthetic realizations generated by the GAN model.The specific surface area $S_V$ and mean chord lengths $\overline{l}_C$ are derived from the slope of the covariance at the origin [see Equ.~(\ref{equ:specific_surface_area})-(\ref{equ:characteristic_pore_size})].}
%\vspace{\floatsep}
\includegraphics[keepaspectratio=True, width=\columnwidth]{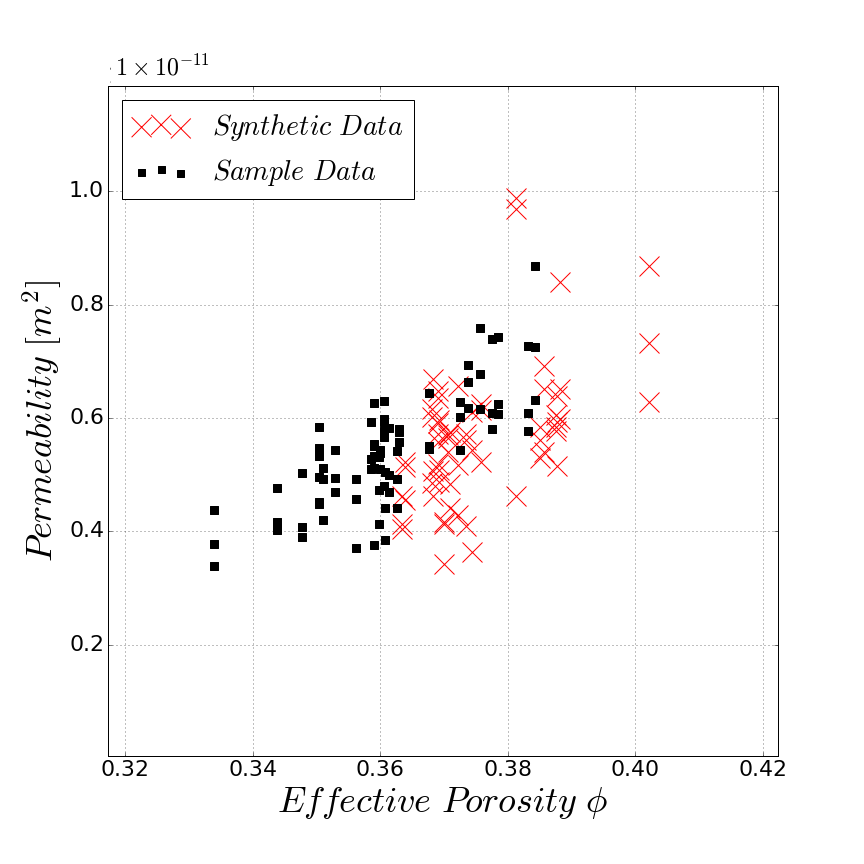}
\caption{\label{fig:permeability_beadpack} A comparison of the numerical estimated single-phase permeability of the beadpack for $128^3$ voxel subdomains of the original image and equal sized GAN based realizations shows an overestimation of the effective porosity for the synthetic models. The mean and variance of both permeability distributions have been found to be in close agreement (see Fig.~\ref{fig:minkowski_beadpack}).}
\end{figure}

The evaluation of the non-centered covariance $S_2(r)$ for the beadpack (Fig.~\ref{fig:radial_correlation_beadpack}) shows a strong hole effect reflecting the spherical nature of the grains. 

\begin{figure*}
\includegraphics[keepaspectratio=True, width=\textwidth]{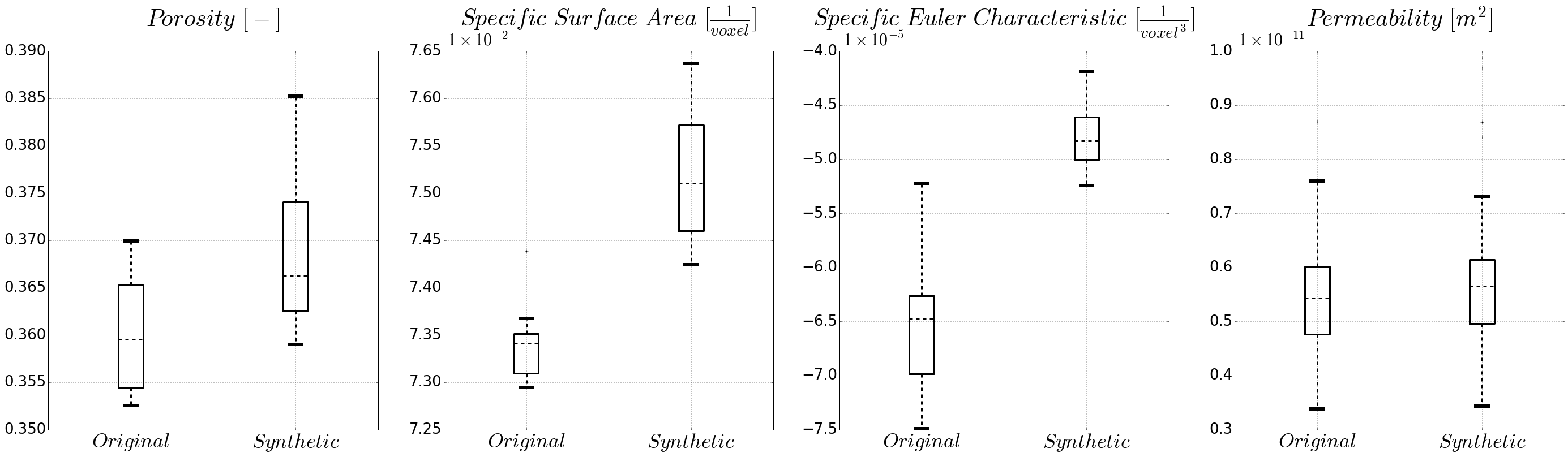}
\caption{\label{fig:minkowski_beadpack} Comparison of three Minkowski functionals for the beadpack evaluated on $200^3$ voxel subdomains of the original training image and realizations of the GAN model. An error of less than $5\%$ can be observed for the porosity and surface area.}
\vspace*{\floatsep}
\includegraphics[keepaspectratio=True, width=\textwidth]{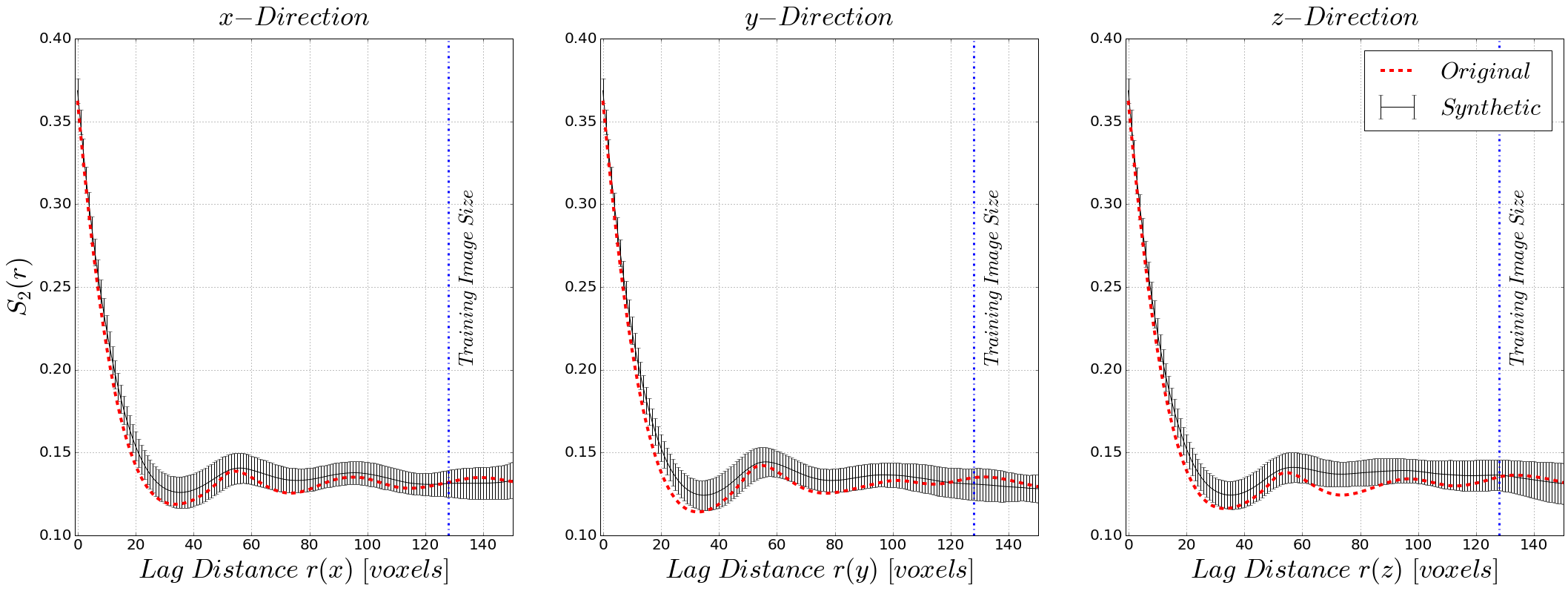}
\caption{\label{fig:correlation_function_beadpack} Comparison of the directional covariance of the beadpack and the average covariance of GAN based synthetic realizations. A clear hole effect can be observed in the original dataset, clearly captured by the GAN model.}

\end{figure*}

A GAN model was trained for 24 hours on the beadpack training image dataset.
The GAN model achieves a small error in the porosity of the generated images with a tendency towards higher porosities (Fig.~\ref{fig:minkowski_beadpack}). \newpage A bias can be observed for the specific surface area and the Euler characteristic of the microstructure (Table~\ref{tab:pore_size_comparison}). 

This bias can be explained by the deviation of the grains from a perfect spherical shape in the synthetic realizations. Due to the smooth nature of the spherical particles in the training image, any deviation from this geometry will lead to an increase in the surface area. This is reflected by a higher specific surface area for the synthetic realizations. In addition we observe a reduction in connectivity, represented by a less negative Euler characteristic.

The directional covariance $S_2$ measured on the generated samples show excellent agreement up to the training image size of $128^3$ voxels and stabilizes at $\phi^2$ (see Fig.~\ref{fig:correlation_function_beadpack}). As expected no directional variation of the covariance is observed and the sample is therefore assumed to be isotropic. 

Single-phase permeability shows a close agreement in both magnitude and variance between the measured training image and the synthetic realizations (Fig.~\ref{fig:minkowski_beadpack}). Figure~\ref{fig:permeability_beadpack} shows a crossplot of the effective porosity $\phi_{eff}$ i.e. the porosity open to flow [Eq.~(\ref{equ:effective_porosity})], and the single-phase permeability exhibiting a similar trend in the distribution of values computed on training images and synthetic realizations. 

\begin{figure*}
\includegraphics[keepaspectratio=True, width=\textwidth]{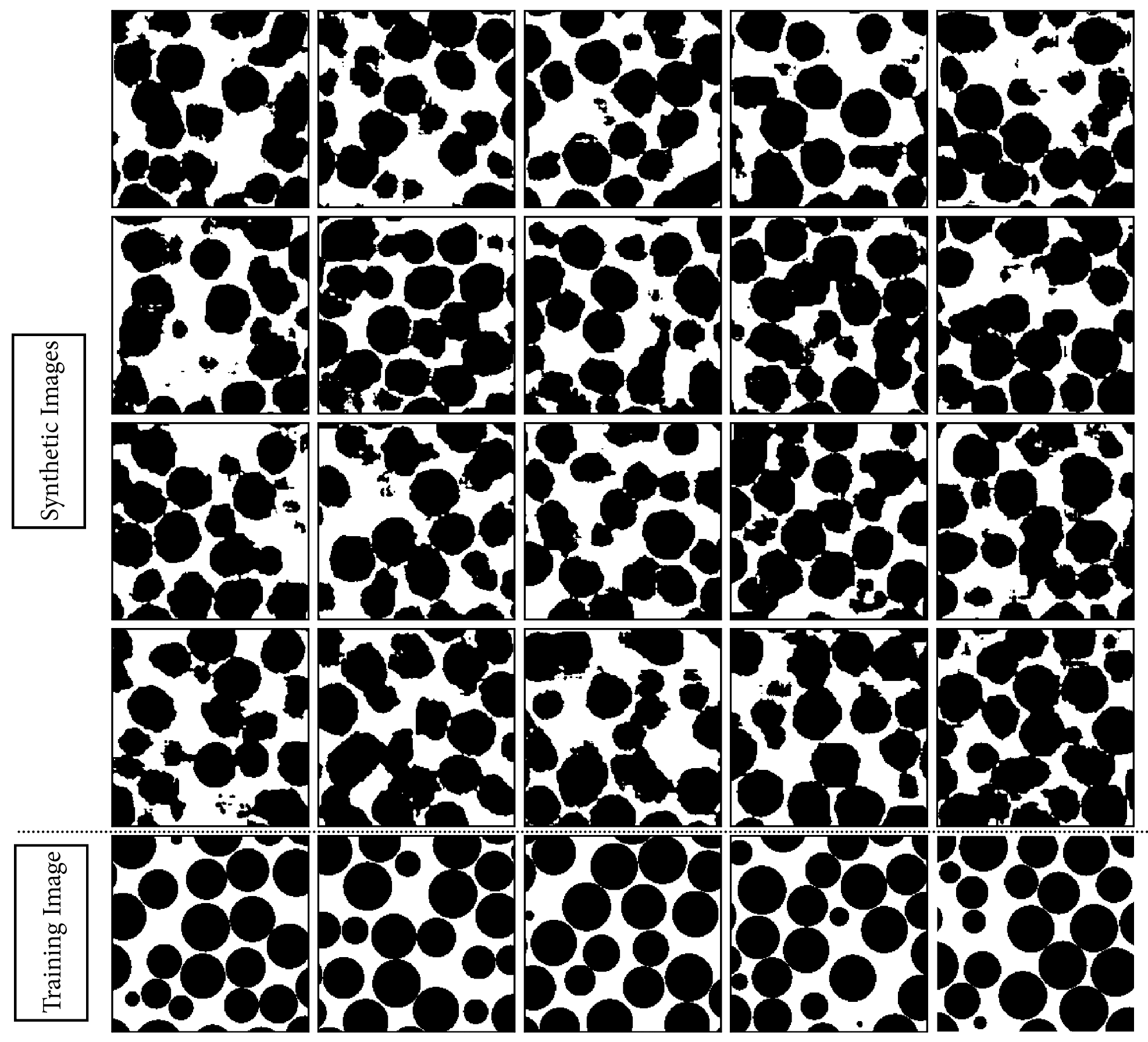}
\caption{\label{fig:beadpack_comparison} Twenty realizations of the spherical beadpack (top) generated to evaluate the statistical, image morphological and transport properties considered in this study. Cross-sectional view of the beadpack training image dataset (bottom).}
\end{figure*}

We provide a comparison of all twenty realizations generated by the GAN model in cross-sections through the x-y plane of the original model and a synthetic realization in Fig.~\ref{fig:beadpack_comparison}. 

Many of the grains show a circular to ellipsoidal shape, which considering the fact that a priori the GAN model does not have any knowledge of the geometry of the grains, learning a representation of a perfect sphere can be considered challenging (see Sec.~\ref{sec:discussion}). The complex grain-grain interface where individual beads contact at single points can be observed for numerous grain arrangements in the generated realizations. 

\clearpage
\subsection{\label{sec:results_berea}Berea}
\begin{figure}
\includegraphics[keepaspectratio=True, width=\columnwidth]{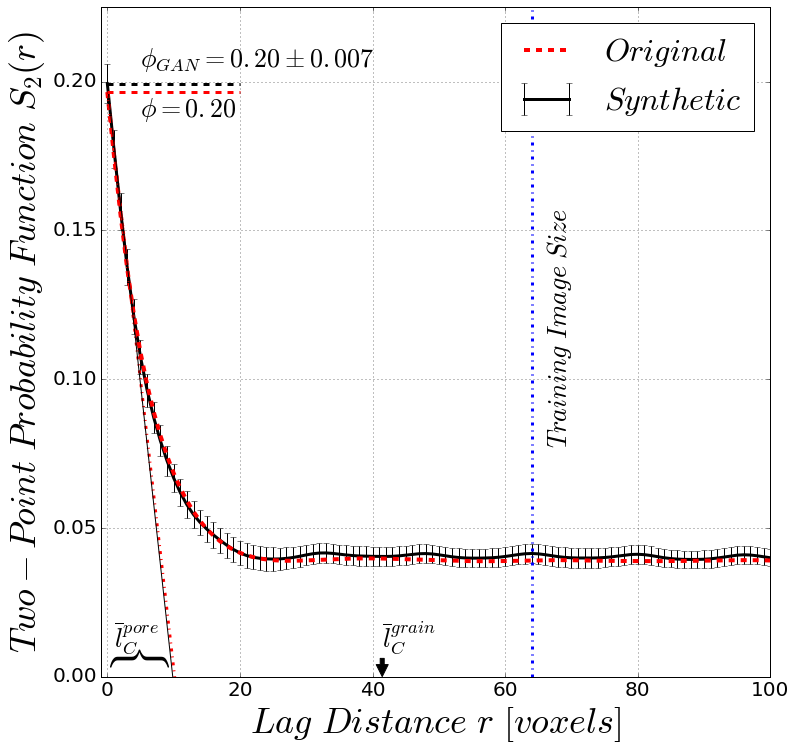}
\caption{\label{fig:radial_correlation_berea} Radial averaged covariance $S_2(r)$ for Berea sandstone training images and 20 synthetic realizations generated by the GAN model.}
\vspace*{\floatsep}
\includegraphics[keepaspectratio=True, width=\columnwidth]{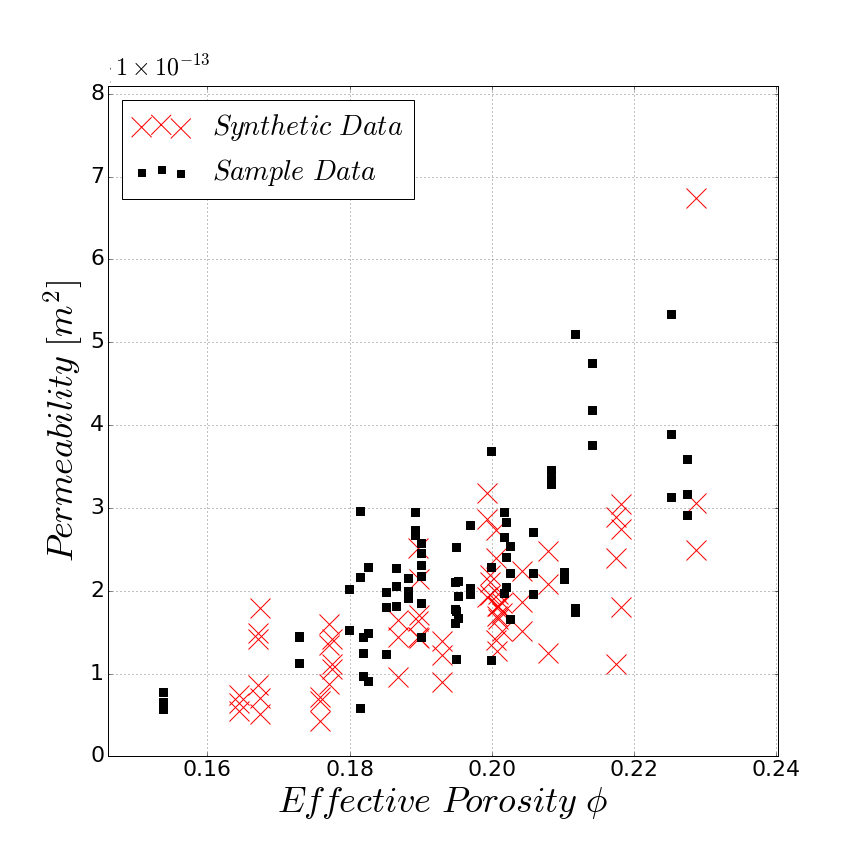}
\caption{\label{fig:permeability_berea} The distribution of numerically obtained permeability values on $128^3$ voxel subdomains and sampled realizations obtained from a GAN model trained on the Berea sandstone dataset show close agreement in the effective porosity, as well as the evaluated permeability.}
\end{figure}

 The radial averaged covariance $S_2(r)$ in Fig.~\ref{fig:radial_correlation_berea}, shows a near exponential decay and stabilisation occurs at a lag distance of 30 voxels for both covariance functions obtained from the Berea training image and synthetic realizations generated by the GAN model.
 
\begin{figure*}
\includegraphics[keepaspectratio=True, width=\textwidth]{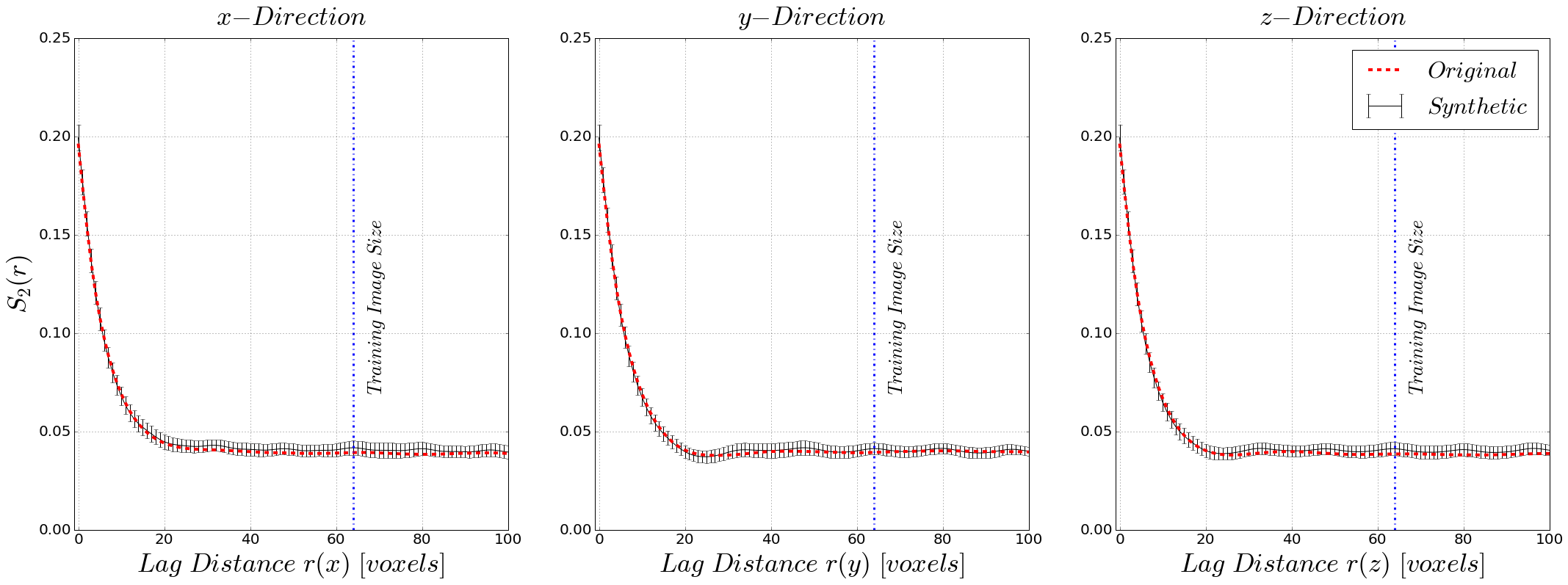}
\caption{\label{fig:correlation_function_berea} Directional non-centered covariance comparison for Berea sandstone. The trained GAN model shows good agreement with the non-centered covariance $S_2$ of the training image.}
\vspace*{\floatsep}
\includegraphics[keepaspectratio=True, width=\textwidth]{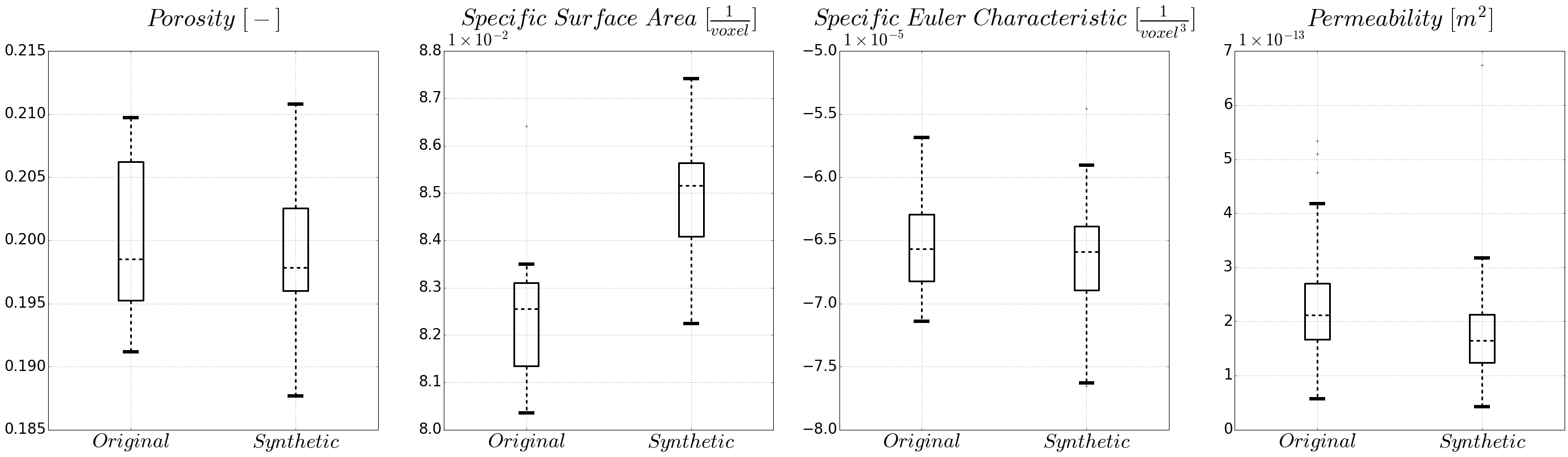}
\caption{\label{fig:minkowski_berea} Comparison of three Minkowski functionals for Berea sandstone. The porosity, specific surface area and specific Euler characteristic show good agreement between the training image and samples from the trained GAN model.}
\end{figure*}

Additionally, Fig.~\ref{fig:correlation_function_berea} shows that the directional two-point statistics characterized by the directional covariances is captured in the generated images. This is shown by comparing the small hole effect observed in the z-direction of the Berea sample with the x-direction where a near exponential decay can be observed. In both cases, the GAN model shows excellent agreement and closely follows the trend of the empirical estimates of $S_2$. 

The results of the direct computation of the Minkowski functionals is presented in Fig.~\ref{fig:minkowski_berea} and show comparable distributions for the porosity $\phi$, specific surface area $S_V$ and the Euler characteristic $\chi_V$ of the training images and the synthetic realizations.

A comparison of the specific surface area $S_V$ obtained from the covariance and the direct computation of the Minkowski functional, show nearly equal values (Table~\ref{tab:pore_size_comparison}).

The obtained estimates of the single-phase permeability show a similar distribution covering the range of effective permeability measured on the training images. Figure~\ref{fig:permeability_berea} shows the computed values of permeability and the corresponding effective porosity. The permeability of the synthetic realizations capture the values, variability and trend obtained from the Berea training image dataset.

\begin{figure*}
\includegraphics[keepaspectratio=True, width=\textwidth]{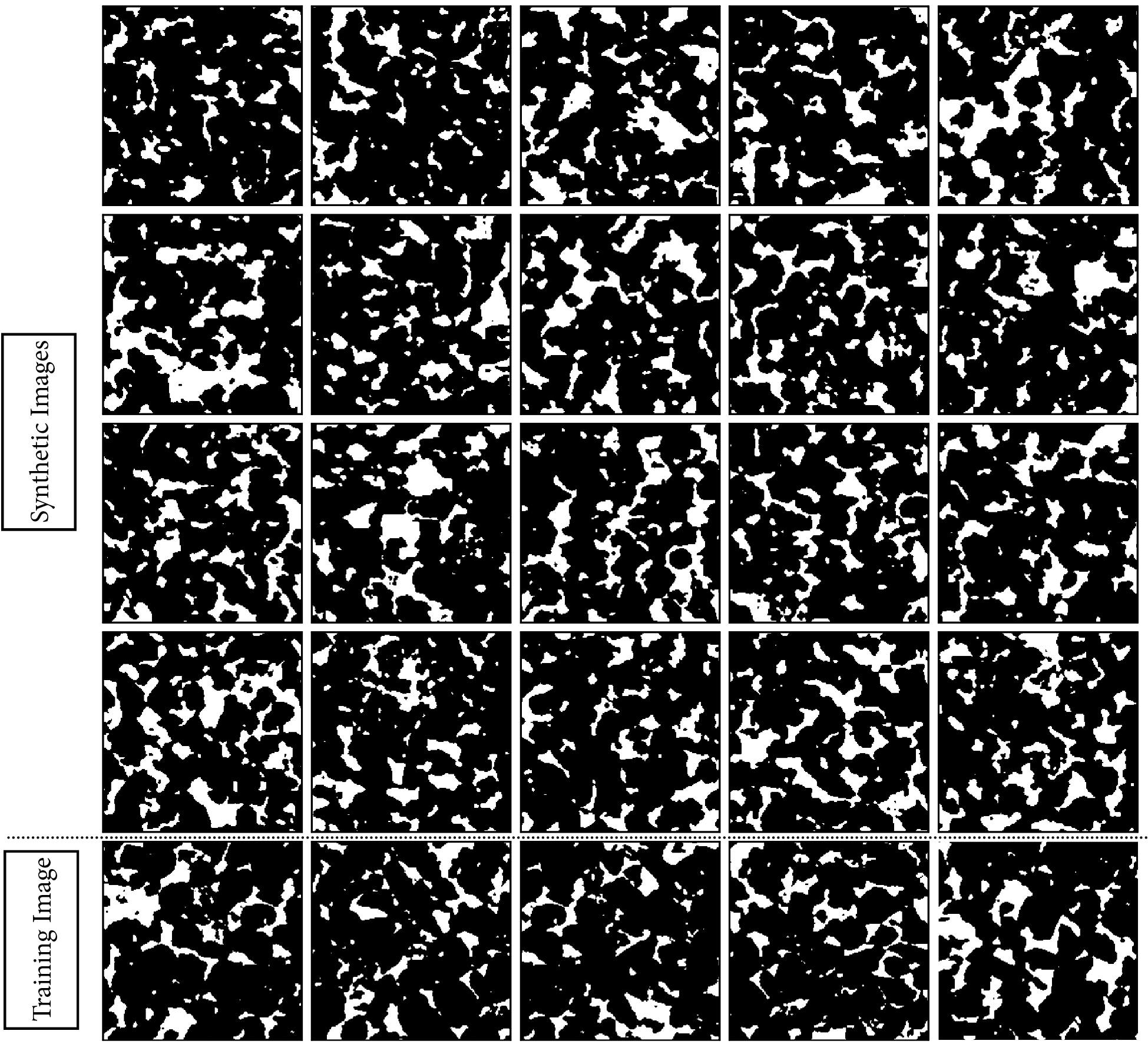}
\caption{\label{fig:berea_comparison} Realizations generated by the GAN model (top) compared to training images (bottom) for Berea sandstone.}
\end{figure*}

Figure~\ref{fig:berea_comparison} shows a comparison of twenty realizations of the GAN model trained on the Berea dataset. A smaller training image size of $64^3$ voxels was used, as compared to the beadpack ($128^3$ voxels). This is due to the smaller size of the structuring elements observed in the training image. A smaller training image size was therefore sufficient to capture the long and short range correlation found in the Berea sample.

\clearpage
\subsection{\label{sec:results_ketton}Ketton}
\begin{figure}
\includegraphics[keepaspectratio=True, width=\columnwidth]{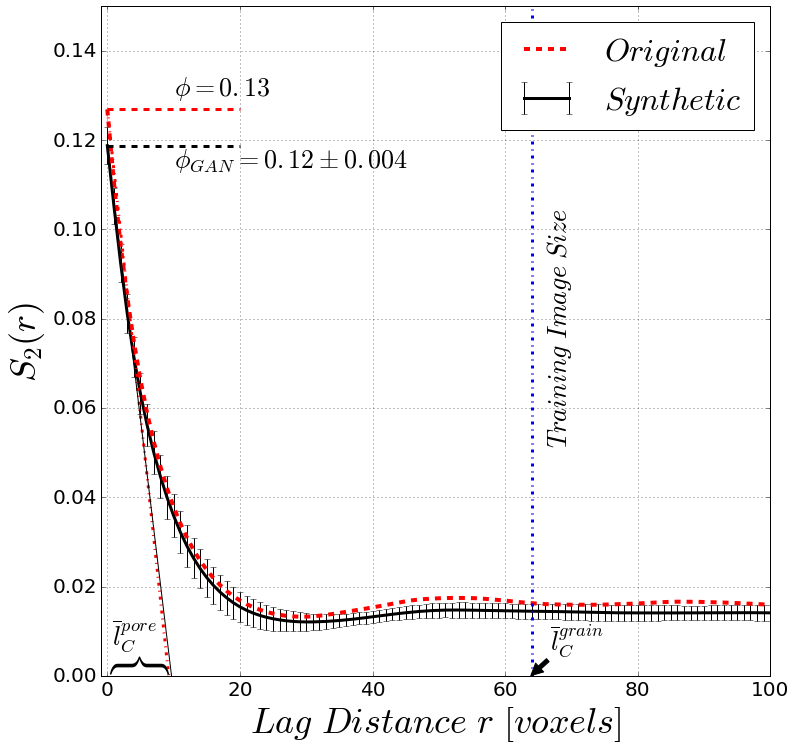}
\caption{\label{fig:radial_correlation_ketton} Radial averaged covariance $S_2(r)$ for the oolitic Ketton limestone training image and 20 synthetic realizations generated by the GAN model.}

\vspace*{\floatsep}

\includegraphics[keepaspectratio=True, width=\columnwidth]{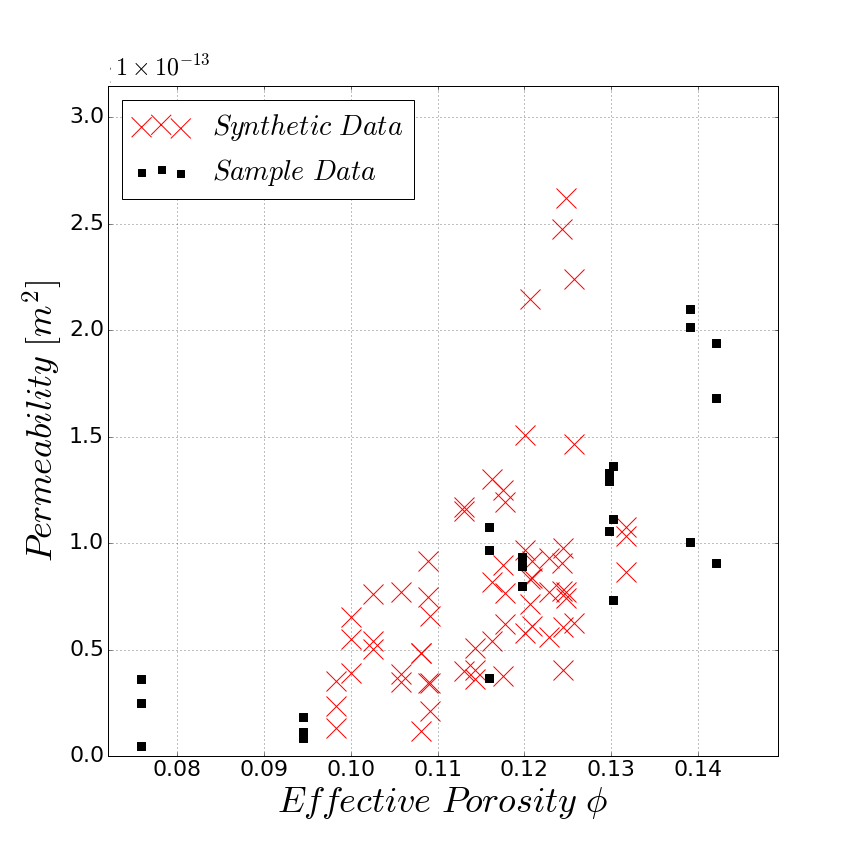}
\caption{\label{fig:permeability_ketton} Evaluated single-phase permeability for the Ketton training image. The synthetic realizations show similar effective porosity and permeability as the Ketton sample.}
\end{figure}

\begin{figure*}
\includegraphics[keepaspectratio=True, width=\textwidth]{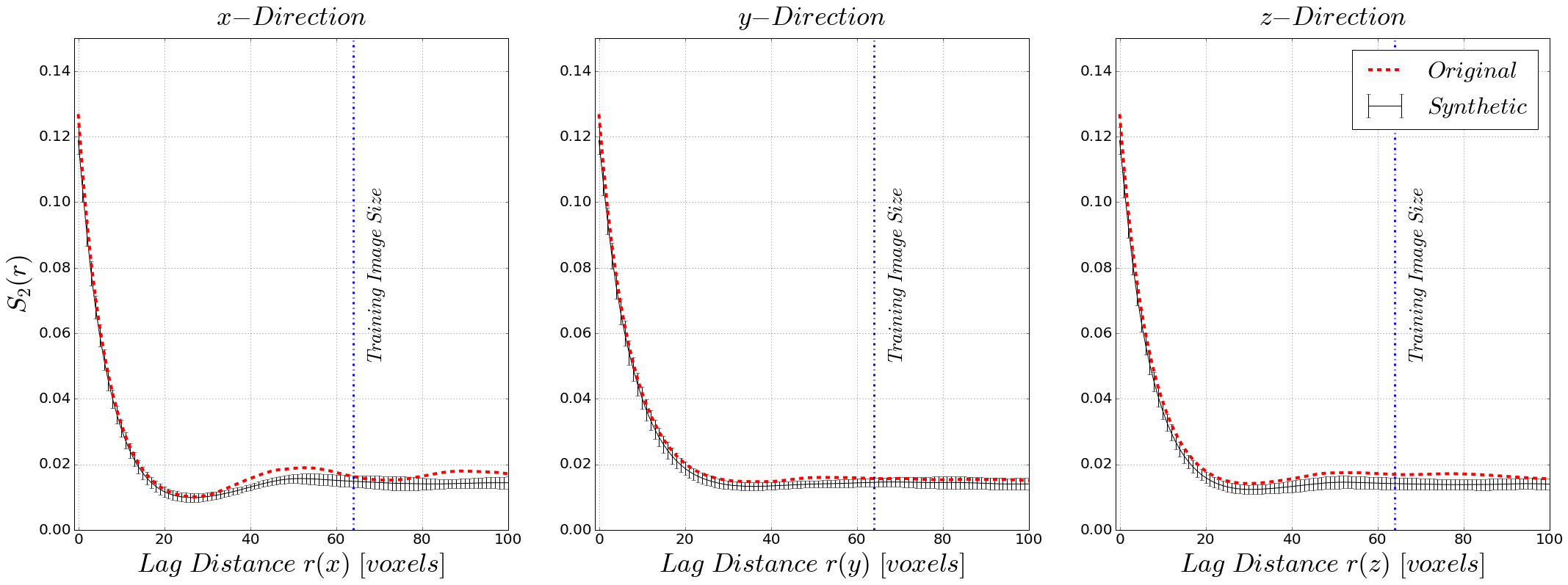}
\caption{\label{fig:correlation_function_ketton} The directional covariance of the Ketton sample shows oscillating behavior in the x-direction, whereas a nearly exponential decrease can be observed for the y and z directions. This anisotropy in $S_2(r)$ is also reflected in the covariance of the samples obtained from the GAN model.}

\vspace*{\floatsep}

\includegraphics[keepaspectratio=True, width=\textwidth]{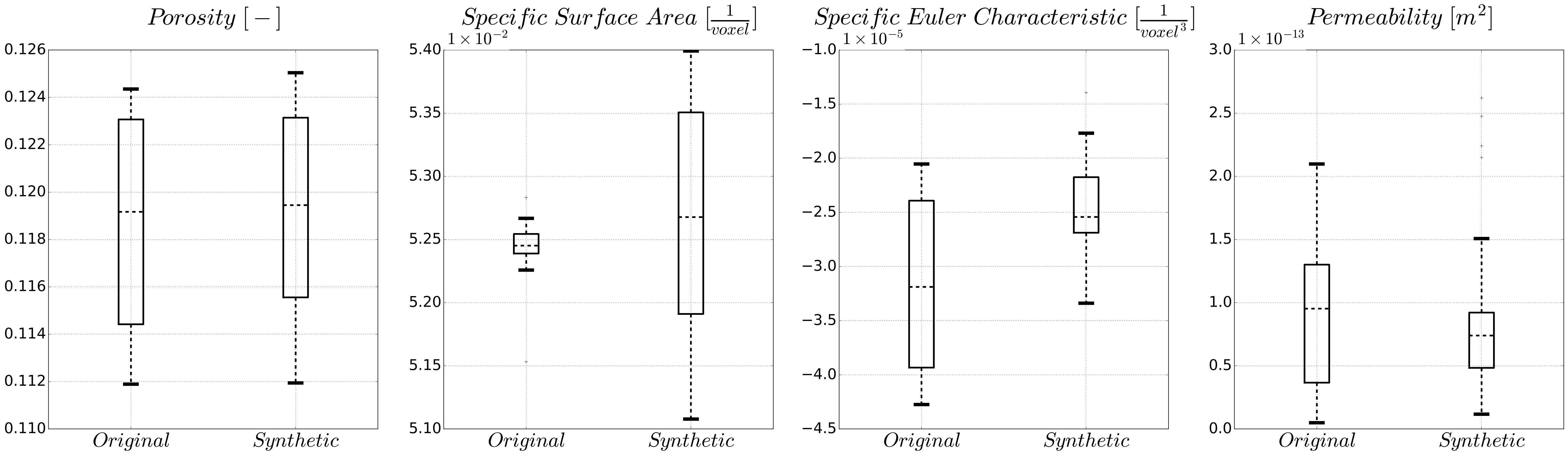}
\caption{\label{fig:minkowski_ketton} Comparison of the Minkowski functionals for the Ketton training image. The three evaluated Minkowski functionals show good agreement. The evaluated Euler characteristic indicates that the sampled synthetic realizations show a similar degree of connectivity as the training image.}
\end{figure*}

The covariance $S_2(r)$ of the Ketton limestone shown in Fig.~\ref{fig:radial_correlation_ketton}, shows a pronounced hole effect due to the ellipsoidal oolitic grains.  Due to the hole effect observed in the radial averaged covariance (Fig.~\ref{fig:radial_correlation_ketton}), we relate the Ketton sample to a hard-sphere model. Figure~\ref{fig:correlation_function_ketton} indicates that the images generated by the GAN model trained on the Ketton image, capture the oscillatory and anisotropic behavior of the covariance observed in Ketton. The specific surface area $S_V$ derived from the generated images is in close agreement with the training data. An error of approximately $1\%$ was achieved in the porosity of the GAN generated images compared to the original Ketton dataset (Fig.~\ref{fig:minkowski_ketton}).

The measured specific surface area of the synthetic images shows a higher variance compared to the original training images. Nevertheless, the average values of the porosity $\phi$ and specific surface area $S_V$ derived from the non-centered covariance $S_2(r)$ [see Eq.~(\ref{equ:specific_surface_area})] are in good agreement with values obtained from direct image morphological estimation  (see Table~\ref{tab:pore_size_comparison}).

The distribution of single-phase permeability estimates of the synthetic GAN realizations overlies the permeability values of the Ketton training images. 

The Euler characteristic $\chi_V$ and the permeability of the Ketton training dataset are closely matched by the synthetic images and therefore capture the connectivity observed in the oolitic Ketton limestone.

\begin{figure*}
\includegraphics[keepaspectratio=True, width=\textwidth]{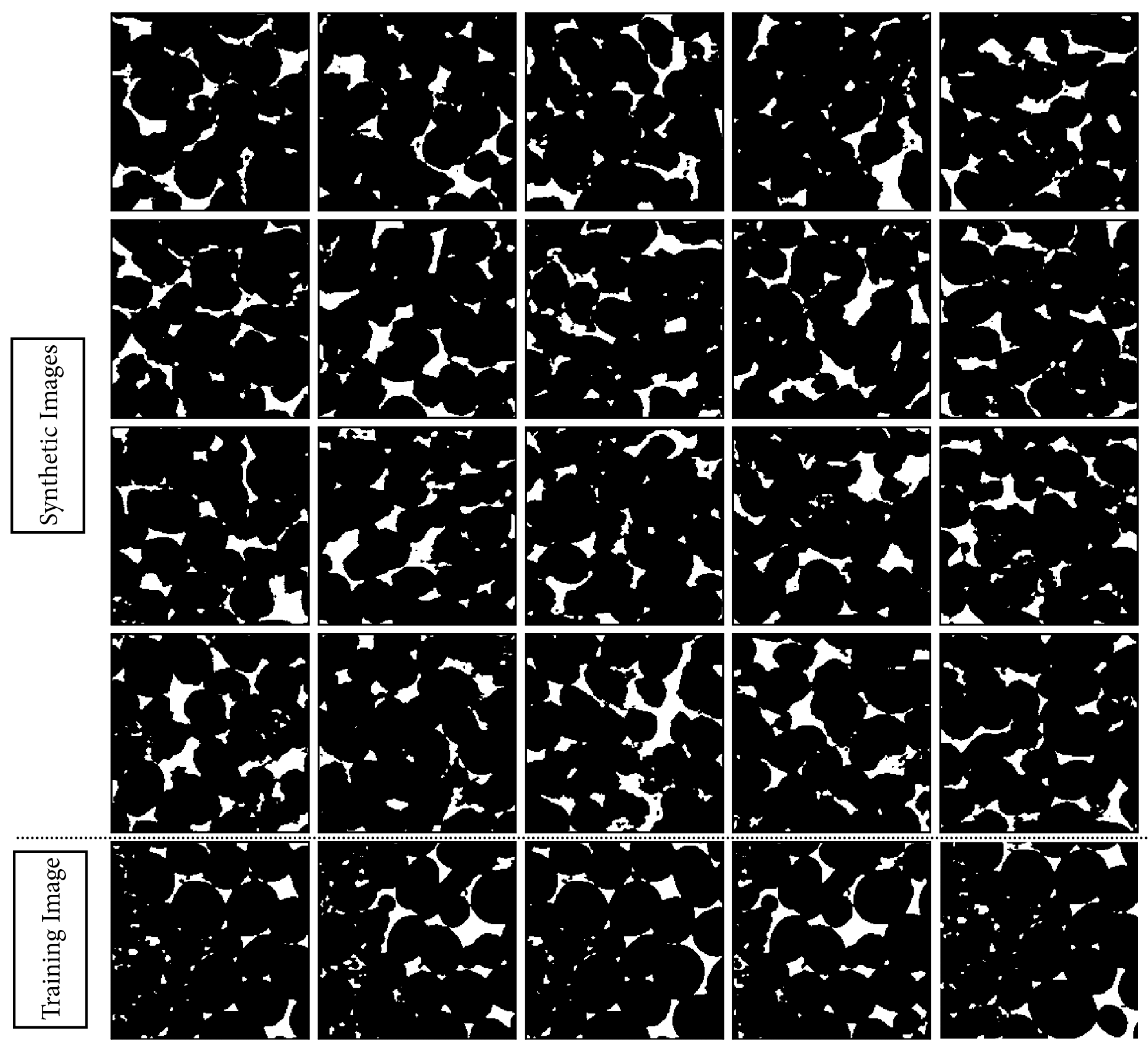}
\caption{\label{fig:ketton_comparison}Realizations generated by the GAN model (top) compared to training images (bottom) for Ketton limestone.}
\end{figure*}

We present an overview of the 20 realizations generated by the GAN model trained on the Ketton dataset in Fig.~\ref{fig:ketton_comparison}.

\clearpage

\section{\label{sec:discussion} Discussion}
This paper presents a novel method for three-dimensional stochastic image reconstruction based on generative adversarial neural networks (GAN) trained on three-dimensional segmented images. To summarize, the objectives of this contribution are threefold. Firstly, the generation of stochastic reconstructions of porous media such as sedimentary rocks exceeding the size of the acquired image datasets. Secondly, to evaluate the ability of GAN models to capture the image morphological and physical properties of micro-scale porous media. Thirdly, to establish a method of stochastic image reconstruction that allows a probabilistic treatment of pore-scale properties such as permeability without the need to acquire numerous images of a single rock type.

The first objective stems from technical limitations of micro-CT data acquisition. Images are acquired as a trade-off between sample size i.e. how many representative structures can be captured in one image versus the resolution at which these pore-scale structures are resolved. The generation of large porous domains based on high-resolution images enables this gap in scales to be bridged and micro-scale features to be incorporated in macro-scale models. 

Our findings show that GANs can learn an implicit representation of the image space given a limited number of training images subsampled from larger images. These subdomains were extracted based on characteristic length scales (see Sec.~ \ref{sec:two_point_statistics}) and serve as a training set for the GAN model. For the Ketton limestone, a small spacing of the extracted subdomains was required to increase the size of the training image dataset. While we did not find any evidence of an introduced bias by using correlated subdomains, we believe that these extracted training images should represent independent regions.

We have evaluated the ability to train GANs for a number of training image sizes less than and up to twice the size of the structuring elements. We have found that models trained on images smaller than the average grain size results in artifacts and distorted shapes occurring in the generated micro-structures. For the beadpack, the size of an individual sphere is 50 voxels. A training image of $64^3$ voxels would typically only contain parts of an individual grain and only capture the interaction of the particles, but not the geometry of the structuring element. For the beadpack, models trained on $64^3$ voxels were successful in learning a representation of the short scale micro-structure but failed to reproduce the long distance correlation. A larger training image of $128^3$ voxels, as was used to model the beadpack has a much higher chance to represent the full geometry of the particles and therefore not only learn interactions, but also the shapes of grains. 

We, therefore, suggest that training images extracted from large datasets must be larger than the average grain size. For models that are well described by a Boolean model, the size of the structuring element can be readily estimated from stabilization of the covariance $S_2(r)$. For more complex samples a different measure must be used to estimate the size of the required training image.

The chord length is one additional measure that can be obtained to characterize the grain space of porous media. While we have found that the mean chord length of the grain space $\overline{l}_C^{grain}$ is always less than or equal to our chosen training image size, $\overline{l}_C^{grain}$ increases with decreasing porosity. This contradicts the need to have the largest training domain for the beadpack sample which also has the highest porosity. A better estimate may be related to the representative elementary volume of the specific surface area which by definition is the same for the grain and pore space and is, therefore, more representative of the morphology of the porous medium \cite{bear2013dynamics}. 
Based on the properties we have evaluated we could not find a measure derived from two-point statistical or image morphological properties that is closely related to the required training image size and we see a theoretical discussion of this as possible future work.

Conceptually the simplest model considered in this study, the spherical beadpack, has proven to be the most challenging as a training image for the GAN model (Sec.~\ref{sec:results_beadpack}). While we observe spherical and ellipsoidal shapes in the resulting realizations (see Fig.~\ref{fig:beadpack_comparison}), the shape is exactly defined by the spherical nature of the grains. Any deviation from this shape, which for GANs, is learned implicitly from the data itself, will lead to a misrepresentation of the effective properties. Random hard-sphere models with spherical grains will efficiently capture the nature of this dataset. Therefore we suggest a fit-for-purpose application of GANs, for training images that exhibit variability of grain sizes and shapes, which are not readily captured by a simpler model.

While for many sedimentary granular rocks representative volumetric images can be obtained, this may be more challenging for carbonate samples with complex pore-grain structures. The three training images considered in this study were all treated under the assumption of stationarity i.e. we do not expect a variation in the mean and variance of the averaged properties as a function of location. In theory, GANs are not limited to learning representations of stationary datasets. This is shown by the many successful applications for two-dimensional image and texture synthesis of non-stationary domains, such as learned image representations of human faces \cite{gauthier2014conditional} or galaxies \cite{Schawinski2017, Ravanbakhsh2016}. Therefore a model that incorporates non-stationarity for a single rock-type would technically be possible in the GAN framework but would require the acquisition of many images of the same porous medium.

A valid representation of the microscale variability and connectivity of the pore space is critical to assess the single and multi-phase flow behavior of porous media. Therefore any stochastic reconstruction method used in the process of deriving or evaluating the variability of micro-scale properties must capture the statistical and image morphological characteristics of the reconstructed porous medium. While we have shown that for the evaluated datasets, the GAN based image reconstructions capture the variation and characteristics of these porous media, a number of challenges arise in this task that are fundamentally different to classical stochastic methods of image reconstruction. 

For porous media, many flow related properties can be related to the porosity. Classical stochastic methods are able to capture the porosity efficiently by defining a specific proportion of the grain and pore domain. The GAN based model presented in this study initially has no knowledge of the porosity. The porosity, therefore, arises as a feature of the training image data. Matching the porosity distribution of the training image dataset was found to be the main challenge in training a GAN model. An error of three percent in porosity would lead to a mismatch in the permeability of the synthetic images. It is, therefore, necessary to continuously monitor the derived properties such as the Minkowski functionals or estimates of the permeability, in the course of training the neural networks to ensure that synthetic realizations created by the GAN model are able to capture the effective properties of the micro-scale domains. 

While this can be considered one of the main challenges in the application of GANs for synthetic image reconstruction, learning an implicit representation of the training data itself can be seen as a strength. Many classic stochastic methods rely on the formulation of an objective function that ensures that statistical properties are captured in the generated realizations e.g. matching $S_2(r)$ and the specific surface area $S_V$ of the stochastic reconstructions to a desired precision. The GAN approach does not require an explicit objective function a priori. The objective function is encoded in the discriminator and adapted in the course of training. 

During adversarial training both the generator and discriminator are continuously improved. The discriminator's sole purpose is to be able to distinguish real training data from generated synthetic data. On the other hand, the generator tries to generate synthetic data that the discriminator is not able to distinguish from the training data. Due to the multi-scale representation of the convolutional neural networks, these features must be learned across the full range of length scales present in the training data, leading to a high-resolution image that captures small and large scale features of the image dataset. A number of stacked GAN models can be trained on e.g. low-resolution medical-CT data and high-resolution micro-CT allowing incorporation of spatial information across multiple length scales \cite{2016arXiv161203242Z}.
 \begin{figure}
\includegraphics[keepaspectratio=True, width=\columnwidth]{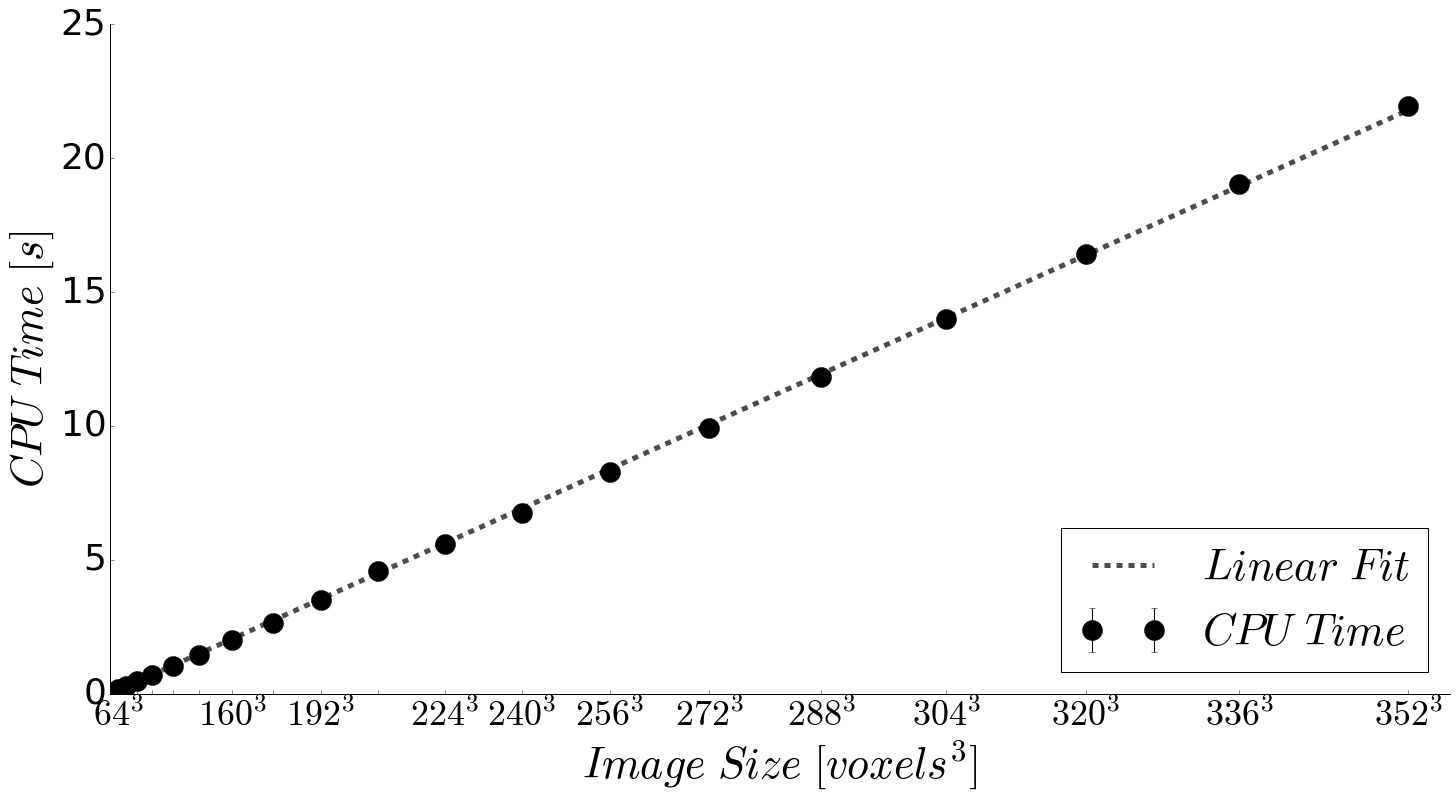}
\caption{\label{fig:timing_scaling_cpu} Measured CPU time for generating synthetic realizations of Berea sandstone at increasing image size. 100 realizations were computed at each image dimension and CPU time averaged. Computational cost increases linearly with the number of voxels in the generated image.}
\end{figure}

Once the GAN model has successfully learned to create physically representative samples of the porous medium, one possible application is to evaluate the variability in the flow properties by evaluating the properties of a large number of samples. This not only requires a physically valid representation of the porous medium but also requires a method that allows fast image reconstruction. In Sec.~\ref{sec:results} we have shown that training was performed for approximately 24 hours and may vary due to the need for manual inspection of the generated samples in the training process. Figure~\ref{fig:timing_scaling_cpu} shows the CPU time required for generation of images at increasing image size. The fully convolutional nature of the GAN architecture allows very large images, exceeding the size of the original sample to be generated very efficiently and at low computational cost and runtime. 

While training requires considerable time and computational resources in the form of modern graphics processors as well as optimized neural network frameworks, image reconstruction requires little computational effort and scales linearly in the total number of voxels of the generated images. This, therefore, enables the generation of ensembles of large domains based on volumetric images acquired from 3D microscopy, that capture the physical behavior of the porous medium. The learned representation of the generator consists of the weights of the convolutional filters learned in the training process and can, therefore, be stored for future use once training has finished.
\section{\label{sec:conclusion}Conclusions}
We have evaluated the application of generative adversarial neural networks (GAN) for stochastic image reconstruction of porous media based on previously acquired images of sedimentary rocks. Three image datasets were used as training images: a beadpack, a Berea sandstone, and an oolitic Ketton limestone.

By evaluating two-point statistical measures, image morphological features and computing the single-phase effective permeability we have shown that the synthetic images generated by the GAN model are able to capture the characteristic statistical and physical behavior of these porous media. \newpage While a large computational effort is required to train the GAN model, the generation of samples from the learned representation is highly efficient and learned models are easily stored for future use.

Future work in the application of GANs to stochastic image reconstruction of porous media will include improving the quality of the image reconstruction by evaluating various generator-discriminator architectures, the use of grayscale and multi-channel training images, as well as the application of large multi-scale domains of porous media to evaluate the ensemble behavior of single and multi-phase flow properties in porous media. Recent advances in the understanding of GANs should lead to a more stable and consistent training process \cite{Mao2016, arjovsky2017wasserstein}.
\begin{acknowledgments}
O. Dubrule thanks Total for seconding him as a \\ Visiting Professor at Imperial College.
\end{acknowledgments}
\bibliography{paper.bib}

\end{document}